\documentclass[acmsmall,nonacm]{acmart}

\usepackage{multirow}
\usepackage{soul}
\usepackage{comment}

\usepackage{amsmath,amsfonts,bm}

\def\eqref#1{Eq.~(\ref{#1})}

\def\1{\bm{1}}

\DeclareMathAlphabet{\mathsfit}{\encodingdefault}{\sfdefault}{m}{sl}
\SetMathAlphabet{\mathsfit}{bold}{\encodingdefault}{\sfdefault}{bx}{n}

\def\gC{{\mathcal{C}}}

\AtBeginDocument{%
  }

\setcopyright{none}
\copyrightyear{2025}
\acmYear{2025}
\acmDOI{XXXXXXX.XXXXXXX}

\def\runningfoot{\def\@runningfoot{}}
\def\firstfoot{\def\@firstfoot{}}
\renewcommand\footnotetextcopyrightpermission[1]{}

\begin{document}

\title{Multi-Agent Collaboration Mechanisms: A Survey of LLMs}

\author{Khanh-Tung Tran}
\email{123128577@umail.ucc.ie}
\orcid{0000-0001-6796-8911}
\affiliation{%
  \institution{School of Computer Science and Information Technology, University College Cork}
  \city{Cork}
  \country{Ireland}}

\author{Dung Dao}
\affiliation{%
  \institution{School of Computer Science and Information Technology, University College Cork}
  \city{Cork}
  \country{Ireland}}
\email{123122658@umail.ucc.ie}

\author{Minh-Duong Nguyen}
\affiliation{%
  \institution{Department of Information Convergence Engineering, Pusan National University}
  \city{Busan}
  \country{South Korea}
}
\email{duongnm@pusan.ac.kr}

\author{Quoc-Viet Pham}
\affiliation{%
  \institution{School of Computer Science and Statistics, Trinity College Dublin}
  \city{Dublin 2, D02PN40}
  \country{Ireland}}
\email{viet.pham@tcd.ie}

\author{Barry O'Sullivan}
\affiliation{%
  \institution{School of Computer Science and Information Technology, University College Cork}
  \city{Cork}
  \country{Ireland}}
\email{b.osullivan@cs.ucc.ie}

\author{Hoang D. Nguyen}
\authornote{Corresponding author.}
\affiliation{%
  \institution{School of Computer Science and Information Technology, University College Cork}
  \city{Cork}
  \country{Ireland}}
\email{hn@cs.ucc.ie}

\renewcommand{\shortauthors}{Tran et al.}

\begin{abstract}


    With recent advances in Large Language Models (LLMs), Agentic AI has become phenomenal in real-world applications, moving toward multiple LLM-based agents to perceive, learn, reason, and act collaboratively.
    These LLM-based Multi-Agent Systems (MASs) enable groups of intelligent agents to coordinate and solve complex tasks collectively at scale, transitioning from isolated models to collaboration-centric approaches. This work provides an extensive survey of the collaborative aspect of MASs and introduces an extensible framework to guide future research. Our framework characterizes collaboration mechanisms based on key dimensions: actors (agents involved), types (e.g., cooperation, competition, or coopetition), structures (e.g., peer-to-peer, centralized, or distributed), strategies (e.g., role-based or model-based), and coordination protocols. 
    Through a review of existing methodologies, our findings serve as a foundation for demystifying and advancing LLM-based MASs toward more intelligent and collaborative solutions for complex, real-world use cases. In addition, various applications of MASs across diverse domains, including 5G/6G networks, Industry 5.0, question answering, and social and cultural settings, are also investigated, demonstrating their wider adoption and broader impacts. Finally, we identify key lessons learned, open challenges, and potential research directions of MASs towards artificial collective intelligence.
\end{abstract}






\maketitle

\hfill \break
\textbf{Reference:} Khanh-Tung Tran, Dung Dao, Minh-Duong Nguyen, Quoc-Viet Pham, Barry O’Sullivan, and Hoang D. Nguyen. 2025. Multi-Agent Collaboration Mechanisms: A Survey of LLMs. arXiv preprint (2025), 35 pages.

\newpage
\section{Introduction} \label{sec:Introduction}
    \subsection{Motivation}
    
        Recent advancements in Large Language Models (LLMs) have transformed artificial intelligence (AI), enabling them to perform sophisticated tasks such as creative writing, reasoning, and decision-making, arguably comparable to human level~\cite{zhao2023survey}. While these models have shown remarkable capabilities individually, they still suffer from intrinsic limitations such as hallucination~\cite{huang2023survey}, auto-regressive nature (e.g., incapable of slow-thinking~\cite{hagendorff2023human}), and scaling laws~\cite{kaplan2020scaling,10.5555/3600270.3602446}. To address these challenges, agentic AI leverages LLMs as the brain, or the orchestrator, integrating them with external tools and agenda such as planning, enabling LLM-based agents to take actions, solve complex problems, and learn and interact with external environments\footnote{\url{https://blogs.nvidia.com/blog/what-is-agentic-ai/}}\textsuperscript{,}\footnote{\url{https://www.ibm.com/think/insights/agentic-ai}}. Furthermore, researchers are increasingly exploring horizontal scaling — leveraging multiple LLM-based agents to work together collaboratively towards collective intelligence. This approach aligns with ongoing research in Multi-Agent Systems (MASs) and collaborative AI, which focus on enabling groups of intelligent agents to coordinate, share knowledge, and solve problems collectively. The convergence of these fields has given rise to LLM-based MASs, which harness the collective intelligence of multiple LLMs to tackle complex, multi-step challenges~\cite{sun2024llm}. Inspiration for MASs extends beyond technological advancements and finds roots in human collective intelligence (e.g., society of mind~\cite{minsky1988society}, theory of mind~\cite{frith2005theory}). Human societies excel in leveraging teamwork and specialization to achieve shared goals, from everyday tasks to scientific discoveries. Similarly, MASs are designed to emulate these principles, enabling AI agents to collaborate effectively by combining their individual strengths and perspectives.

\begin{figure}
            \centering
            \includegraphics[width=0.9\linewidth]{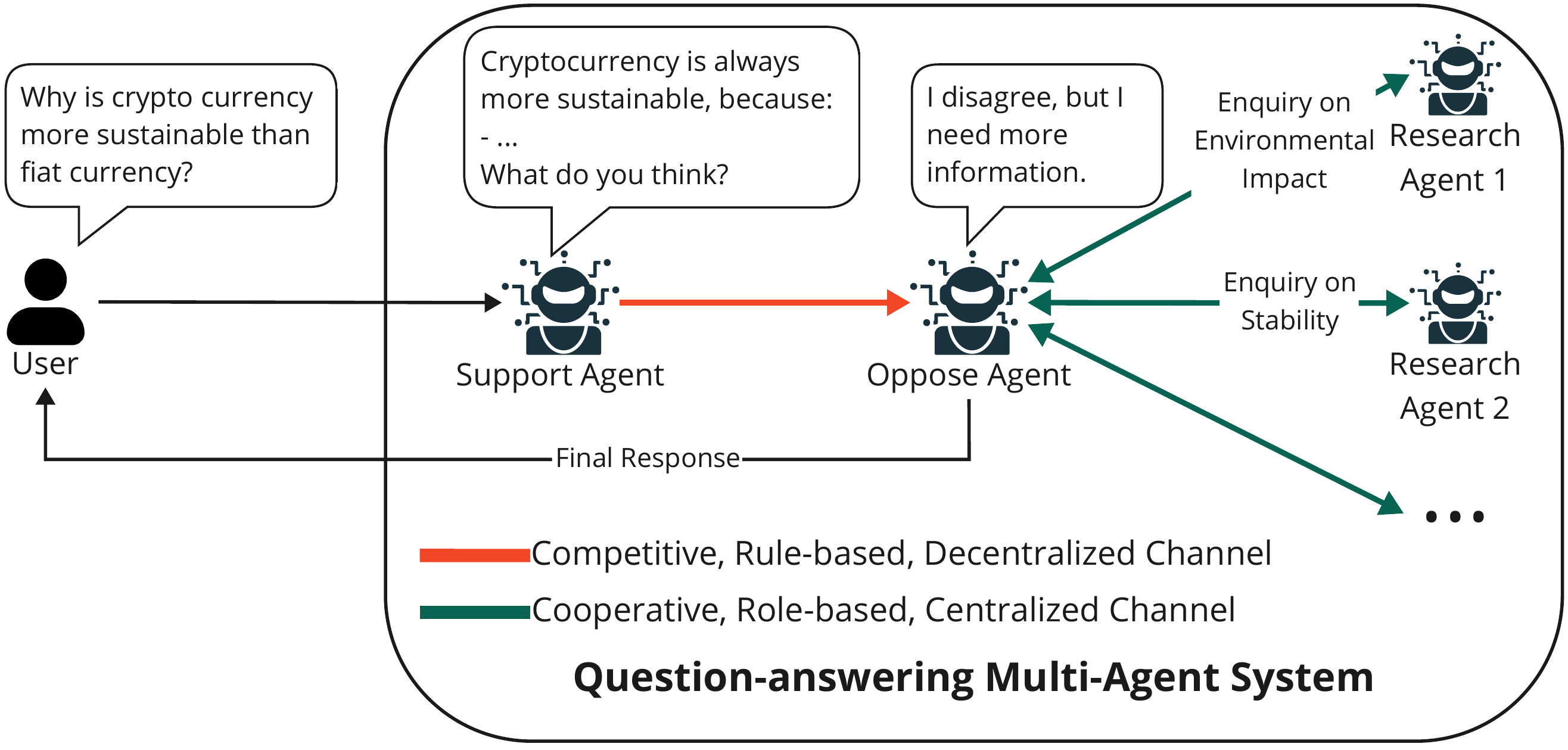}
            \caption{Example question-answering application of LLM-based multi-agent collaborative system. In the first collaboration channel, two LLMs are collaborating through a debate against each other, given the input by the user with a turn-based strategy. In the second channel, the Oppose Agent cooperates and leverages information from Research Agents, and provides the final response to the user.}\label{fig:example}
        \end{figure}

        LLM-based MAS can have multiple collaboration channels with different characteristics, as illustrated in Fig.~\ref{fig:example}.
        MASs have demonstrated notable successes across various domains, enhancing the capabilities of individual LLMs by leveraging collaboration and coordination among specialized agents. These systems distribute tasks among agents, allowing agents to share knowledge, execute subtasks, and align their efforts toward shared objectives. The potential benefits of MASs are transformative. They excel in knowledge memorization, enabling distributed agents to retain and share diverse knowledge bases without overloading a single system~\cite{hatalis2023memory,zhang2024survey}. They enhance long-term planning by delegating tasks across agents, supporting persistent problem-solving over extended interactions~\cite{huang2024understanding}. Furthermore, MASs enable effective generalization by pooling expertise from multiple models with specialized prompts/personas, allowing them to address diverse problems more effectively than standalone models. Lastly, MASs improve interaction efficiency by simultaneously managing subtasks through specialized agents, accelerating the resolution of complex, multi-step tasks. MAS strives to achieve collective intelligence, where the combined capabilities of multiple agents exceed the sum of their individual contributions~\cite{chen2024agentverse}.

        One of the main focus for effective MASs is the mechanisms of collaboration~\cite{li-etal-2023-theory,li2023camel,das2023enabling,pan2024agentcoordvisuallyexploringcoordination,wang-etal-2024-unleashing}, which lead to a transition from traditional, isolated models toward approaches that emphasize interactions, enabling agents to connect, negotiate, make decisions, plan, and act jointly, driving forward the capabilities of AI in collective settings. A deeper understanding of how collaboration mechanisms operate in MASs is critical to unlocking their full potential.
    \\
    \\

    \subsection{State-of-the-Arts and Contributions}

        Due to the importance and timely need for LLM-based multi-agent collaborative systems, there have been a couple of surveys on this topic. However, these works often fall short in fully addressing the collaborative aspects and mechanisms of LLM-based MASs, which are crucial to enabling agents to work effectively toward shared goals, as summarized in Table~\ref{tab:related_surveys}. For instance,~\cite{xi2023risepotentiallargelanguage,ijcai2024p890,10.1145/3704435} focus on single-agent systems and only touch on multi-agent collaboration at a surface level. \cite{xi2023risepotentiallargelanguage} lays the groundwork by proposing a framework for LLM-based agents, consisting of three components: brain, perception, and action. Their work highlights the use of LLMs as the brain of agents, leveraging techniques such as input modality integration, prompting, retrieval, and tool usage. However, their discussion of multi-agent collaboration is limited to agent behaviors and personalities, lacking an exploration of mechanisms that enable agents to work together. \cite{ijcai2024p890} surveys the domains and settings where LLM-based MASs have been successfully applied, profiling the communication structures of these systems (layered, decentralized, centralized, and shared message pools) but without touching other characteristics of collaboration, such as type, strategy, or coordination architecture. 
        
        Other works, such as~\cite{lu2024mergeensemblecooperatesurvey}, focus on collaborative strategies, categorizing them into merging, ensemble, and cooperation. Although their survey discusses how these strategies are applied to LLMs and extends cooperation beyond traditional fusion techniques, it overlooks other essential collaboration mechanisms, such as competition and coopetition, and dimensions beyond popular collaboration types. In contrast,~\cite{talebirad2023multiagentcollaborationharnessingpower} proposes a generic framework for enhancing LLM capabilities via MASs, showing how tools like Auto-GPT align with their framework. However, the collaboration mechanisms remain conceptual, lacking detailed implementation and characterization. In~\cite{han2024llmmultiagentsystemschallenges}, the focus is on configuring LLMs to leverage diverse capabilities and roles, such as integrating memory and information retrieval components. Their exploration of multi-agent collaboration primarily centers on planning and orchestration architectures, emphasizing global and local task planning based on agent roles and specializations. Meanwhile,~\cite{Gao2024} narrows its focus to the application of LLM-based MASs in agent-based modeling and simulation, discussing challenges such as environment perception, human alignment, action generation, and evaluation. While insightful for simulation-specific applications, it lacks a broader perspective on in-depth collaborative mechanisms. Similarly,~\cite{10.1145/3697350} surveys these systems for digital twin applications, while~\cite{he2024llm,10.1145/3704806} focuses on the domain of software engineering.

        \begin{table*}
          \caption{Summary of related surveys on LLM-based multi-agent collaborative system.}
          \label{tab:related_surveys}
          \resizebox{1.0\linewidth}{!}{
          \begin{tabular}{ccccc}
            \toprule
            \multirow{3}{*}{\shortstack[c]{\textbf{Refs.}}} & \multirow{3}{*}{\shortstack[c]{\textbf{Focus on Multi-} \\ \textbf{Agent Collaborative} \\ \textbf{System}}} & \multirow{3}{*}{\shortstack[c]{\textbf{Review of Collaborative}  \\ \textbf{Aspects and Mechanisms} \\ \textbf{in MAS}}} & \multirow{3}{*}{\shortstack[c]{\textbf{Propose General} \\ \textbf{Framework for} \\ \textbf{MAS}}} & \multirow{3}{*}{\shortstack[c]{\textbf{Review of Real-}\\\textbf{World Applications}}} \\ \\ \\
            \midrule
            \cite{xi2023risepotentiallargelanguage} & {\color{OrangeRed}Low}  & {\color{OrangeRed}Low} & {\color{BrickRed}None} & {\color{BrickRed}None} \\
            \cite{10.1145/3704806} & {\color{OrangeRed}Low} & {\color{OrangeRed}Low} & {\color{BrickRed}None} & {\color{OrangeRed}Low} \\
            \cite{lu2024mergeensemblecooperatesurvey} & {\color{Orange}Medium}  & {\color{OrangeRed}Low} & {\color{BrickRed}None} & {\color{BrickRed}None} \\
            \cite{han2024llmmultiagentsystemschallenges} & {\color{Orange}Medium} & {\color{OrangeRed}Low} & {\color{BrickRed}None} & {\color{OrangeRed}Low} \\
            \cite{10.1145/3697350} & {\color{Orange}Medium} & {\color{OrangeRed}Low} & {\color{BrickRed}None} & {\color{OrangeRed}Low} \\
            \cite{talebirad2023multiagentcollaborationharnessingpower} & {\color{Orange}Medium} & {\color{BrickRed}None} & {\color{OrangeRed}Low} & {\color{Orange}Medium} \\
            \cite{Gao2024} & {\color{Orange}Medium} & {\color{OrangeRed}Low} & {\color{BrickRed}None} & {\color{Orange}Medium} \\
            \cite{ijcai2024p890} & {\color{Orange}Medium}  & {\color{OrangeRed}Low}  & {\color{Orange}Medium} & {\color{ForestGreen}High}\\
            \hline
            OURS  & {\color{ForestGreen}High} & {\color{ForestGreen}High} & {\color{ForestGreen}High} & {\color{ForestGreen}High} \\
          \bottomrule
          \end{tabular}
          }
        \end{table*}

        From the summary and explanation above, there are clear gaps in fully exploring the collaborative aspects and mechanisms of LLM-based MASs, which are crucial for enabling agents to work together toward shared goals. This work aims to provide a comprehensive view of the collaborative foundations between LLM-based agents in multi-agent collaborative systems. With collaboration as the main focus, our study characterizes collaborations between agents based on their actors (agents involved), type (e.g., cooperation, competition, or coopetition), structure (e.g., peer-to-peer, centralized, or distributed), and strategy (e.g., role-based, rule-based, or model-based), and the coordination layer in collaborations. We emphasize the mechanisms and know-how that facilitate effective collaboration, identifying key characteristics and trends in MAS design. Through a survey of existing approaches and identification of open challenges, we synthesize these findings into a cohesive framework. This framework serves as a foundation for future research, advancing the integration of LLMs in MASs and paving the way for more adaptable, intelligent, and cooperative AI systems capable of addressing complex, real-world applications.
		
        Our main contributions are listed as follows:
        \begin{itemize}
            \item \textbf{Collaborative Aspects and Mechanisms in LLM-based MAS}: we focus on the operational mechanisms of LLM-based multi-agent collaboration, emphasizing the "know-how" required to enable effective collaboration, including the collaboration type, strategy, communication structure and coordination architecture.
            \item \textbf{General Framework for LLM-based MAS}: we present a comprehensive framework, integrating diverse characteristics of MAS, allowing researchers to understand, design and develop multi-agent collaborative systems.
            \item \textbf{Review of Real-World Applications}: we examine real-world implementations of LLM-based MASs across various domains, highlighting their practical applications, successes, and limitations.
            \item \textbf{Discussion of Lessons Learned and Open Problems}: we identify key challenges in the developmental agenda of MASs, such as collective reasoning and decision-making, and outline potential research directions to address these challenges.
        \end{itemize}

    \subsection{Paper Organization}
        This paper is organized as follows. Section~\ref{sec:Background} provides the background necessary for understanding this work, including an overview of LLMs, MASs, and collaborative AI. In Section~\ref{sec:Prelim}, we introduce foundational concepts in LLM-based multi-agent collaborative systems through mathematical notations, emphasizing the vital role of collaboration. Then, in Section~\ref{sec:Method}, we present an extensive review of LLM-based multi-agent collaborative systems, categorized by key characteristics of collaboration, including type, strategy, structure, and coordination and orchestration. 
        Next, Section~\ref{sec:Applications} reviews key applications of LLM-based multi-agent collaborative systems across both industry and academia. In Section~\ref{sec:OpenProblem}, we discuss open problems and potential future research directions in this relatively new and evolving research area. Finally, we conclude this survey paper on LLM-based multi-agent collaborative system in Section~\ref{sec:Conclusion}. 

\section{Background} \label{sec:Background}
    \subsection{Multi-Agent (AI) Systems}
        MAS is a computerized system composed of multiple interacting intelligent agents. The key components of MAS are as follows:
        \begin{itemize}
            \item Agents: The core actors with roles, capabilities, behaviors and knowledge models. Capabilities like learning, planning, reasoning and decision making lend intelligence to the agents and overall system.
            \item Environment: The external world where agents are situated in and can sense and act upon. Environments can be simulated or physical spaces like factories, roads, power grids etc.
            \item Interactions: Communications between agents happen via standard agent communication languages. Agent interactions involve cooperation, coordination, negotiation and more based on system needs.
            \item Organization: Agents either have hierarchical control or organize based on emergent behaviors.
        \end{itemize}


        MASs can solve problems that are difficult or impossible for an individual agent or a monolithic system to solve~\cite{divband2022intelligent}. Agents collaboratively solve tasks yet they offer more flexibility due to their inherent ability to learn and make autonomous decisions. Agents use their interactions with neighboring agents or with the environment to learn new contexts and actions. Subsequently, agents use their knowledge to decide and perform an action on the environment to solve their assigned tasks~\cite{fischer2021loop}. It is this flexibility that makes MAS suited to solve problems in a variety of disciplines including computer science, civil engineering, and electrical engineering. 

        The salient features of MAS, including flexibility, and reliability, self-organization, and real-time operation make it an effective solution to solve complex tasks, which can be detailed as follows: 
        \begin{itemize}
            \item Flexibility and Scalability: MAS can flexibly adapt to changing environments by adding, removing, and modifying agents. This makes them highly scalable for solving complex problems.
            \item Robustness and Reliability: Decentralization of control leads to continued system operation even with some failed components. This lends greater robustness and fault tolerance.
            \item Self-Organization and Coordination: Agents can self-organize based on emergent behavior rules for the division of labor, coordinated decision making, and conflict resolution.
            \item Real-time Operation: Immediate situational responses are possible without the need for human oversight. Enables applications like disaster rescue and traffic optimization.
        \end{itemize}
        Their efficiency stems from the division of labor inherent in MAS whereby a complex task is divided into multiple smaller tasks, each of which is assigned to a distinct agent. Naturally, the associated overheads, e.g., processing and energy consumption, are amortized across the multiple agents, which often results in a low-cost solution as compared to an approach where the entire complex problem is to be solved by one single powerful entity. Each agent can solve the allocated tasks with any level of pre-defined knowledge which introduces high flexibility. The distributed nature of problem solving adopted in MAS also imparts high reliability. In the event of agent failure, the task can be readily reassigned to other agents.

    \subsection{Large Language Models}
        LLMs - driven by the development of transformer architectures \cite{vaswani2017attention} - represent a significant leap in Natural Language Processing (NLP) and AI. These models, such as OpenAI’s GPT \cite{achiam2023gpt}, Meta’s LLaMA \cite{touvron2023llama}, and Google’s Gemini series \cite{team2023gemini}, are trained on vast text corpora and rely on large-scale artificial neural networks with billions, sometimes trillions, of parameters. Their scale has enabled breakthroughs in language understanding, generation, and task-specific applications~\cite{nguyen2023vigptqa,tran2024uccix,10433480,peikos2024leveraging,dong2024can}.

        The defining characteristic of LLMs is their size and the phenomenon of emergent abilities, which arise when models exceed a certain threshold in terms of parameters. These emergent behaviors allow LLMs to solve tasks they were not explicitly trained on, such as analogical reasoning and zero-shot learning, where the model can tackle new problems without additional fine-tuning~\cite{schaeffer2024emergent}. The launch of models like GPT-3 and ChatGPT in recent years has made these capabilities accessible to the public, leading to a surge in both academic and industrial research on how to optimize, scale, and secure LLMs for real-world use~\cite{fan2024bibliometric}.
    
        Despite the promising innovations, LLMs are not without challenges. Their performance degrades as real-world knowledge changes, prompting a focus on aligning models with up-to-date information without retraining from scratch~\cite{cuconasu2024power,chang2024survey}. Moreover, the geopolitical and ethical implications of LLM development have become the limelight for policymakers, especially concerning the computational power required and potential misuse by malicious actors~\cite{10.1145/3605943,10.1145/3649449}.

        LLMs are increasingly being utilized as the "brain" for individual agents in MASs, bringing sophisticated reasoning and language capabilities to each agent. With frameworks like AgentVerse, LLMs enhance agents' autonomy by allowing them to infer tasks, make decisions based on situational awareness, and even exhibit emergent social behaviors such as collaboration and negotiation~\cite{chen2024agentverse}. While LLMs have shown remarkable performance in single-agent tasks, their limitations become apparent in multi-agent settings where the complexity of coordination, communication, and decision-making is higher. Issues such as cascading hallucinations — where one erroneous output leads to compounding mistakes pose challenges in sustained multi-agent interactions. However, frameworks like MetaGPT introduce meta-programming techniques including structured workflows and processes within agent interactions to decompose and tackle complex problems, mitigating these issues~\cite{hong2024metagpt}. Moreover, consensus-seeking mechanisms like those tested in the Consensus-LLM project show that LLMs can negotiate and align on shared goals in dynamic environments~\cite{Chen2023MultiAgent}. These works showcase LLMs' potential as central decision-making components and highlight LLMs' capacity to adapt to the strategies of other LLM-based agents, which could be foundational in multiple applications.

    \subsection{Collaborative AI}
        Collaborative AI often refers to AI systems designed to work together with other AI agents or humans~\cite{9918176}. Collaborative AI emerges from two primary research directions: 1) the advancements of AI which resulted in increasingly effective tools for human use and a growing demand for AI systems that can collaborate with other agents (humans or AI models), and 2) the realization that active collaboration among AI models can significantly enhance efficiency and effectiveness. This research spans various domains, including MASs, human-machine interaction, game theory, and natural language processing~\cite{abad2017autonomous,dafoe2020openproblemscooperativeai,Conitzer_Oesterheld_2024,braccini2024swarm}. By integrating these technologies, collaborative AI has the potential to drive novel applications with profound economic and societal impacts~\cite{MenaGuacas2023,OUYANG2024100616}.

        Collaboration is the key in enabling AI agents to interact and work with each other. A straightforward collaboration mechanism would be two models cooperate together towards a shared goal. While cooperation is a fundamental aspect, the collaboration spectrum extends further, encompassing advanced mechanisms like competition and coopetition.
        Collaborative AI leads to a transition from traditional, isolated models toward approaches that emphasize collaboration. New methodologies have been proposed to enable agents to interact, negotiate, make decisions, plan, and act jointly, driving forward the capabilities of AI in collective settings~\cite{Dafoe2021}.

        A major topic of Collaborative AI is MAS research, which focuses on the interactions between intelligent agents and emergent collaborative behaviors. More specifically, MASs are interested in agents, or AI models, that can learn, adapt, and collaborate with one another in complex environments, towards a common shared goal. On the other hand, the rapid advancement of LLMs has enabled new possibilities. LLMs have been shown to be capable of serving as the ``brains'' behind agents in MASs, driving applications where agents not only perform tasks but interact with external tools (e.g., internet searches, calculators), and, more significantly, with each other~\cite{ijcai2024p890}. However, LLMs are not inherently designed and trained to communicate with one another, leaving a wide array of potential applications and open problems in this area. The fusion of LLMs and MASs promises exciting opportunities for further exploration and innovation.
        
        This work provides a comprehensive view of the collaborative aspect between LLM-based agents in MASs, emphasizing the mechanisms that enable agents to work effectively toward shared goals. By surveying existing approaches and identifying open challenges in this emerging research area, we offer a unique perspective that extends beyond traditional cooperation to include diverse modes of collaboration, such as debate, negotiation, and competition. This in-depth focus on collaborative dynamics positions our work as an essential resource for advancing the integration of LLMs in MASs, paving the way for more adaptable, intelligent, and collaborative AI systems with enhanced capabilities for real-world applications.

\section{Multi-Agent Collaboration Concept} \label{sec:Prelim}
    We introduce the main concepts of LLM-based multi-agent collaborative systems, defining key components of agents, systems, and collaboration mechanisms based on insights from recent research in this emerging area.

    \subsection{Agent and Collaborative System Definition}
        \label{sec:Agent_system_def}
        An agent can be mathematically represented by $a=\{ m, o, e, x, y \}$ as follows:
        \begin{itemize}
            \item Model $m=\{\textrm{arch}, \textrm{mem}, \textrm{adp}\}$: the AI model, consisting of its architecture ($\textrm{arch}$), agent's specific memory ($\textrm{mem}$), and optional adapters ($\textrm{adp}$). Adapters are adaptive intelligent modules that allow the agent to incorporate additional knowledge from others through mechanisms such as speculative decoding and parameter-efficient adapter, which can further enrich the model’s response capabilities~\cite{8352646,introMAC,leviathan2023fast}. In the case of LLM agents, the architecture $\textrm{arch}$ is a language model, and the agent's specific memory $\textrm{mem}$ is typically the system prompt $r$.
            \item Objective $o$: the objective or goal of the agent, guiding its actions within the system. For example, in question-answering tasks, the objective is to minimize the cross-entropy between the generated answer and the ground truth.
            \item Environment $e$: the environment or context encompassing the state or conditions in which the agent operates. In LLM, usually the context window is constrained by the number of tokens.
            \item Input $x$: input perception, such as text or sensor data. In LLM, $x$ is encoded as a sequence of tokens.
            \item Output $y$: the corresponding action or output, defined by the function $y=m(o,e,x)$, where the agent uses its model, context, and objective to act on the input $x$.
        \end{itemize}
        %
        LLM-based agents are typically pre-trained on diverse datasets to provide a strong foundational knowledge base. This pre-training process equips an individual agent with essential skills and understanding, ensuring they can meaningfully contribute to the collaborative environment. Moreover, each agent can also be equipped with external tools unique to their own, such as Calculator and Python interpreter.

        When generalized to a multi-agent collaborative system $S$, it includes the following:
        \begin{itemize}
            \item $\mathcal{A}=\{a_i\}_{i=1}^n$: LLM-based agents, where $n$ is the number of agents, which is pre-defined or adjusted dynamically depending on the current requirements of the system.
            \item $\mathcal{O}_{collab}$: a collective set of goals that may be partitioned into unique objectives for each agent, ensuring alignment with the overall system goal.
            \item $\mathcal{E}$: a shared environment from which agents receive contextual data. In our case of an LLM-based MAS, the environment may take various forms, such as vector-based databases or common messaging interfaces.
            \item $\mathcal{C}=\{{c}_{j}\}$: a set of collaboration channels that facilitate interactions among agents, enabling the exchange of information based on given objectives, environment, and inputs: $y_j = c_j(\{a_{i} (o_i, \mathcal{E}, x_i) | a_{i}, o_i, x_{i} \in c_j \} )$. Channels are distinguished by their mechanisms, including actors (agents), types, structures, and strategies. If two channels differ in these aspects, they are treated as separate channels. We assume that the agents have a common ground, meaning the interface can be understood clearly among them (e.g., all agents use English and are on-topic). 
            \item $x_{collab}$: the input perceived by the system.
            \item $y_{collab}$: the system’s output, modeled as $y_{collab}=S(\mathcal{O}_{collab},\mathcal{E},x_{collab}|\mathcal{A},\mathcal{C})$.
        \end{itemize}

    Through this structured workflow, agents in an LLM-based MAS can collaborate adaptively, responding to the task requirements, learning from each other, and coordinating actions to achieve shared objectives. An example can be seen in~\cite{dong-etal-2024-villageragent}, where the collaboration channels are pre-defined through a Directed Acyclic Graph with each edge as agents handling and receiving outputs, allowing the agents to effectively collaborate towards performing tasks in a simulated Minecraft game environment. Another instance is illustrated in~\cite{10.1145/3672456}, where the collaboration between agents is planned first, then the agents are carried out to perform the coding task.

\begin{figure}
        \includegraphics[width=0.95\textwidth]{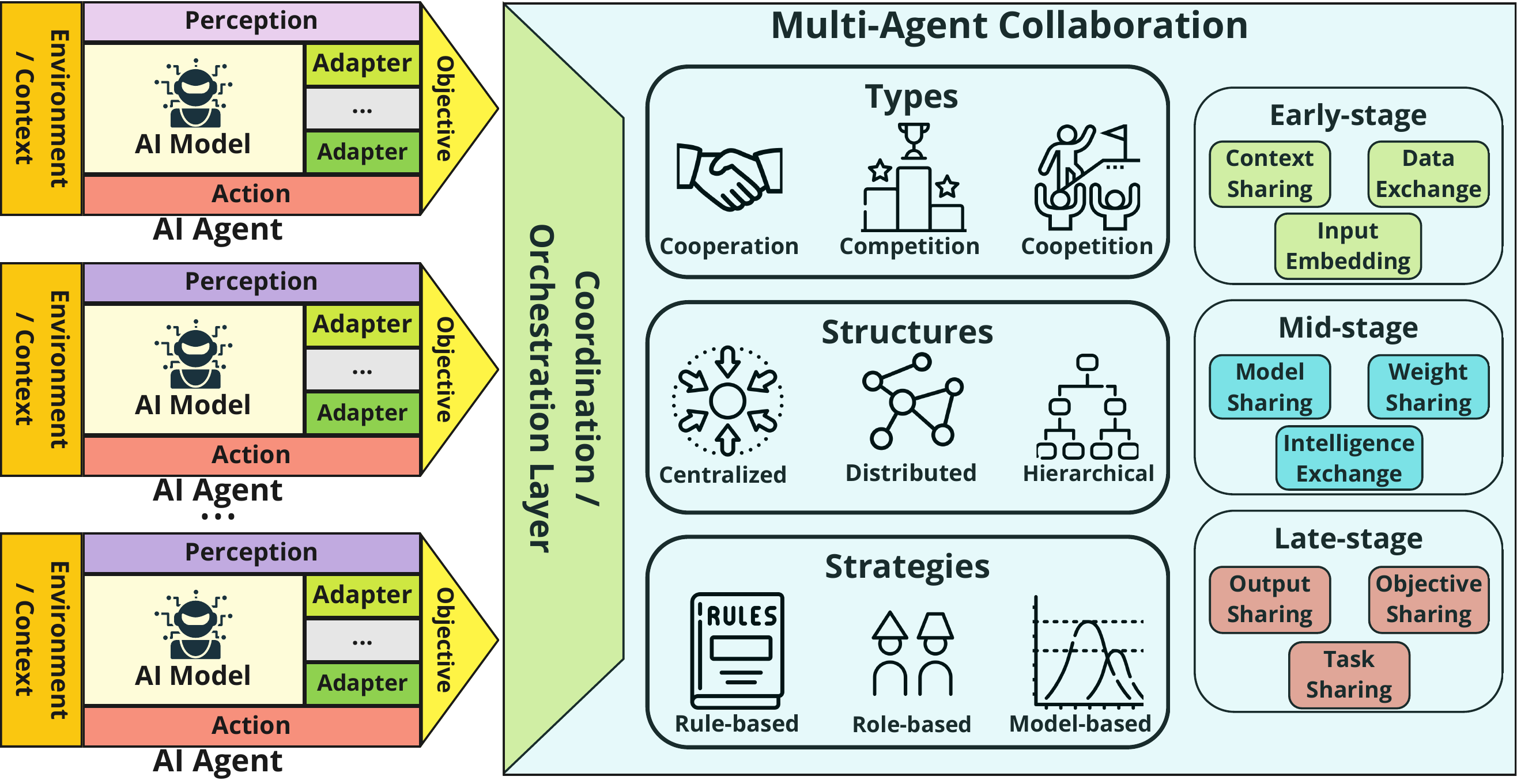}
        \caption{Our proposed framework for LLM-based multi-agent collaborative system. Each agent consists of a language model $m$ as the neural processor, current objective $o$, environment $e$, input perception $x$ and corresponding output/action $y$. The framework's central focus is on collaboration channels $\mathcal{C}$ that facilitate coordination and orchestration among agents. These channels are defined by their actors (agents involved), type, structure, and strategy. Our framework is flexible, accommodating previous approaches and enabling the analysis of diverse MASs under a unified structure.}\label{fig:framework}
        \Description{}
    \end{figure}

    Fig.~\ref{fig:framework} illustrates the agent and its components, as well as the concept of multi-agent collaboration. By defining these components, we can better analyze the collaborative mechanisms necessary for complex, goal-oriented AI collaborations. For instance, a straightforward collaboration mechanism is majority voting, similar to ensemble learning. Collaboration can occur at different stages: (i) late-stage collaborations, such as ensembling outputs/actions $y$ towards collaborative goals, (ii) mid-stage collaborations, for example, exchanging parameters or weights of multiple models $m$ in federated and privacy-preserving manners, and (iii) early-stage collaborations include but not limited to sharing data, context, and environment for model development.

    \subsection{Problem Definition}
        In an LLM-powered MAS, it is vital for the agents to collaborate with each other, sharing a common objective or set of objectives. Each collaboration has a communication channel $c$. The collaboration includes 1) delegating agents (two or more) with certain objectives based on their unique expertise and resources, 2) defining their collaboration mechanisms for working together, and 3) decision making between agents to reach the final goal.
        \begin{equation}
            y_{collab}=S(\mathcal{O}_{collab},\mathcal{E},x_{collab}|\mathcal{A},\mathcal{C})=\{c_j(\{a_{i} (o_i, \mathcal{E}, x_i) | a_{i}, o_i, x_{i} \in c_j \}) | c_j\in \mathcal{C}\}
        \end{equation}
        where each $c_j \in \mathcal{C}$ 
        represents a communication channel facilitating the collaborative actions of agents $a_{i}$ 
        based on their respective inputs $x_{i}$, and allowing them to work together. Working together here goes beyond communication (the exchange of information), requiring deeper collaborative behaviors involving coordination and management, and is key to enabling the capabilities of MASs.

        Each collaboration channel serves as the mechanism through which agents work together, characterized by specific attributes: actors (agents involved), type, structure, and strategy. For instance, channel types can vary, encompassing competition, cooperation, or coopetition, while structures can be peer-to-peer, centralized, or distributed. A difference in any attribute results in a distinct collaboration channel. As an example, in a peer-to-peer structured system, two LLMs may compete, while others cooperate; these distinct interaction types result in separate collaboration channels. This flexible channel framework allows agents to adapt their interactions, optimizing multi-agent collaborative effort and task efficiency across diverse scenarios.

\section{Methodology} \label{sec:Method}


    \subsection{Overview}
        This section provides an extensive review of LLM-based multi-agent collaborative systems, emphasizing their key characteristics, including the mechanisms for coordination and orchestration among agents - collaboration channels - types, strategies, and structures. Fig.~\ref{fig:framework} presents our proposed framework for MASs, detailing their core components and interconnections.

        Our survey strategy involves systematically analyzing existing research on MASs to identify the defining characteristics of multi-agent collaboration. From this analysis, we deduce the fundamental components and trends in MAS design and synthesize them into a cohesive framework. First, each LLM-based agent in the system is equipped with an LLM $m$, current objective $o$, environment $e$, input perception $x$, and corresponding output/action $y$. This is visualized in the left part of Figure~\ref{fig:framework} and described formally using mathematical notations in Section~\ref{sec:Agent_system_def}. Our central focus in this framework is the collaboration channels $\mathcal{C}$ between agents that facilitate coordination and orchestration among agents. These channels are characterized by their actors (agents involved), type (e.g., cooperation, competition, or coopetition), structure (e.g., peer-to-peer, centralized, or distributed), and strategy (e.g., role-based, rule-based, or model-based). 
        Collaboration mechanisms span various levels of machine learning processes, including data exchange, shared input embeddings, model sharing, and output sharing, enabling agents to interact effectively and leverage each other’s strengths.

        For each component, we discuss the prevailing implementation trends and methodologies observed in the literature. We examine how these methods align with our proposed framework. We summarize our main findings and lessons learned at the end of the section, offering guidance for future research in the field.

    \subsection{Collaboration Types}

        \subsubsection{Cooperation}
            Cooperation in LLM-based MASs occurs when agents align their individual objectives ($o_i$) with a shared collective goal ($\mathcal{O}_{collab}$), working together to achieve a mutually beneficial outcome: $\mathcal{O}_{collab} = \bigcup_{i=1}^n o_i$. 
            Agents assess each other’s needs and capabilities, actively seeking collaborative opportunities. Moreover, agents can also be utilized to focus on specific sub-tasks within their expertise, enhancing efficiency and reducing completion times~\cite{chen2024agentverse}. This type of collaboration is essential in tasks where collaborative problem-solving, collective decision making, and complementary skill sets contribute to achieving complex objectives that a single agent could not complete as effectively~\cite{dafoe2020openproblemscooperativeai,das2023enabling,Conitzer_Oesterheld_2024}.

            Several research papers highlight the importance of cooperation in LLM-based MASs. For instance, in~\cite{shinn2023reflexion}, a feedback loop is carried out as the main collaboration channel, where the task is first handled by an LLM model (Actor), then an Evaluator and Self-Reflection model rates the output and results, producing verbal guidance for the Actor to improve. In Theory of Mind for Multi-Agent Collaboration~\cite{li-etal-2023-theory}, agents gain a shared belief state representation within the environment $\mathcal{E}$, helping them track each other’s goals and actions, thereby facilitating smoother coordination and better collaborative outcomes. This shared state has led to emergent collaborative behaviors and high-order Theory of Mind capabilities in LLM agents, though challenges remain in optimizing long-horizon planning and managing hallucinations. In AgentVerse~\cite{chen2024agentverse}, agents specialize in distinct roles, such as recruitment, decision-making, or evaluation, within a cooperative framework, which improves system efficiency by leveraging each agent’s unique expertise. Similarly, MetaGPT~\cite{hong2024metagpt} uses an assembly line model, assigning roles and encoding Standardised Operating Procedures (SOPs) into prompts $r_i$ to enhance structured coordination and produce modular outputs $y_i$. MetaGPT underscores the potential of integrating human domain knowledge into MASs. Cooperative approaches have shown success in areas like question answering~\cite{he-etal-2023-lego}, recommendation systems~\cite{10.1145/3626772.3657669}, and collaborative programming~\cite{islam-etal-2024-mapcoder}, where agents cooperate together with specialized roles, such as manager, searcher, or analyst, to achieve complex goals.

            \begin{table*}
          \caption{Collaboration types: definitions, advantages, disadvantages, and example scenarios.}
          \label{tab:collaboration_types}
          \resizebox{1.0\linewidth}{!}{
          \begin{tabular}{cllllc}
            \toprule
            \textbf{Type} & \multicolumn{1}{c}{\textbf{Definition}} & \multicolumn{1}{c}{\textbf{Advantages}} & \multicolumn{1}{c}{\textbf{Disadavantages}} & \textbf{Example Scenario} & \textbf{Refs.} \\
            \midrule
            Cooperation & Agents align their & \textbullet~Assigns sub-tasks to & \textbullet~Misaligned goals can & Code generation & \cite{shinn2023reflexion,barbarroxa2024benchmarking,hong2024metagpt,islam-etal-2024-mapcoder} \\
             & objectives and work & agents based on strengths. & cause inefficiencies. & Decision making & \cite{nascimento2023self,shinn2023reflexion} \\
            & together toward a & \textbullet~Simple to design and & \textbullet~One agent's failure & Game environments & \cite{li-etal-2023-theory} \\
            & shared goal. &  execute with clear goals. & can be amplified. & Question answering & \cite{li2023camel,shinn2023reflexion,das2023enabling,he-etal-2023-lego} \\
            & & & & Recommendation & \cite{10.1145/3626772.3657669} \\
            \hline
            Competition & Agents prioritize & \textbullet~Pushes agents to perform & \textbullet~Needs mechanisms & Debate & \cite{liang-etal-2024-encouraging} \\
            & their own objectives, &  better. &  to resolve conflicts. & Game environments & \cite{chen-etal-2024-llmarena,zhao2024competeai} \\
            & which may conflict & \textbullet~Promotes adaptive &  \textbullet~Ensures competition  & Question answering & \cite{he-etal-2023-lego,puerto-etal-2023-metaqa} \\
            &  with others. & strategies. & is beneficial. \\
            \hline
            Coopetition & A blend of  & \textbullet~Balances trade-offs to  & \textbullet~Few studies explore  & Negotiation & \cite{davidson2024evaluating,abdelnabi2024cooperation} \\
            & cooperation and & reach mutual agreements. & coopetition in depth. & & \\
            & competition where & & & & \\
            & agents collaborate \\
            & on shared tasks \\
            & while competing & & & & \\
            & on others. \\
          \bottomrule
          \end{tabular}
          }
        \end{table*}

        \begin{figure}
            \includegraphics[width=0.95\textwidth]{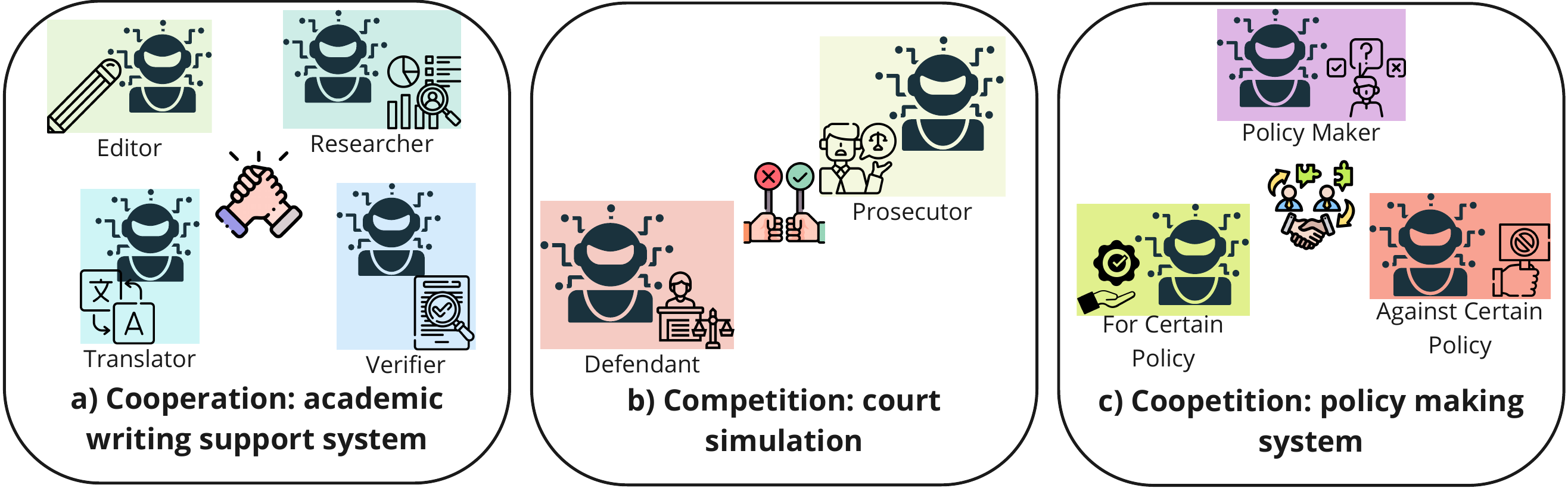}
            \caption{Illustrative examples of collaboration types, where each agent $a$ is equipped with different tools or capabilities through their system prompt $r$. In scenario a), agents cooperate by leveraging their individual specialties (e.g., writing, translation, research) to achieve a shared goal (academic writing). In scenario b), agents compete and debate against each other fo
            r their own goals. In scenario (c) of coopetition, agents compromise with each other, compete on one aspect while agreeing on another. 
            }
            \label{fig:collaboration_types}
            \Description{}
        \end{figure}
            
            There are recent open-source frameworks allowing for experimentation with cooperative LLM-based MASs. CAMEL~\cite{li2023camel} provides a role-playing framework where a task-specific agent and two cooperating AI agents (User and Assistant) work to complete tasks via role-based conversations. Similarly, AutoGen~\cite{wu2024autogen} enables developers to define flexible agent behaviors and communication patterns, allowing LLM agents to cooperate through conversation and tackle complex tasks by decomposing them into manageable subtasks.
            
            However, cooperation in MASs also presents challenges. Frequent communication and multiple collaboration channels in $\mathcal{C}$ between agents can lead to increased computational cost and complexity. Coordinating actions between multiple agents, particularly in dynamic environments, can also be difficult without well-defined collaboration channels $c_i$. Although cooperation is the primary goal, conflicts may arise if agents interpret shared objectives differently or if situations require dynamic adaptation. For example, in the book marketplace application described in~\cite{nascimento2023self}, agents may act unpredictably by sending messages to themselves, pretending to be clients. The overall success of a cooperative MAS is also dependent on the reliability and performance of individual agents, as the failure of one agent or more agents (e.g., infinite conversation loop, amplified hallucinations~\cite{hong2024metagpt}) can negatively impact the entire system. Therefore, mechanisms such as failure handling and trustworthiness need to be considered.

        \subsubsection{Competition}
            Competition happens when there are conflicting objectives or scenarios of limited resources. In this type of interaction, agents prioritize their individual goals ($o_i$), which may clash with or oppose the objectives of others, introducing an element of rivalry: $\mathcal{O}_{collab}=\{o_i | o_i \neq o_j, \forall i \neq j \}$. However, this competition can still orient toward the collective goal $O_{collab}$, such as in the scenario of debate. In LLM-based MASs, competitive dynamics can emerge in tasks such as debate, or strategic gameplay, where agents seek to maximize their own success criteria~\cite{chen-etal-2024-llmarena,zhao2024competeai}.

            Incorporating competition into collaborative MASs can enable innovation and improve the robustness of agents' responses. Competition encourages agents to develop advanced reasoning and more creative problem-solving and strengthens the system's adaptability by testing the limits of each agent’s capabilities. In frameworks like LLMARENA~\cite{chen-etal-2024-llmarena}, LLM-based MASs with competition as the main collaboration type, are benchmarked across seven dynamic gaming environments.  For instance, in the game TicTacToe, the board is represented textually within the environment $\mathcal{E}$, and two LLM agents are instructed (through their system prompts $r_i$) compete, aiming to out-maneuver each other since their individual goals $o_i$ are mutually exclusive. Crucially, the authors highlight that competition between LLM agents enables skills such as spatial reasoning, strategic planning, numerical reasoning, risk assessment, communication, opponent modeling, and team collaboration. However, they also acknowledge that LLMs still have a significant journey ahead in their development towards becoming fully autonomous agents, especially in opponent modeling and team collaboration, due to their intrinsic limited capability to interact with other actors. A game environment is also simulated in~\cite{zhao2024competeai}, where 2 agents act as two restaurant managers competing for 50 customers.  Carefully designed prompts $r$ set the scenario, contextualizing the agents' environment ($\mathcal{E}$) and providing a comprehensive restaurant management system accessible through APIs (external tools). Each agent's context $e_i$ includes information about the rival’s performance from the previous day, including the menu, number of customers, and feedback. In this scenario, the collaboration channel $c$ between the two managers is competitive, illustrating how structured competition drives agents to refine strategies, conforming to several classic sociological and economic theories. Similarly, in LEGO~\cite{he-etal-2023-lego}, a multi-agent collaborative framework is introduced for causality explanation generation, where the competition collaborative link $c$ is also pre-defined. The collaboration consists of 2 LLMs, one serves as Explainer with initial output, and another one acts as Critic, with iterative refinement and feedback. In~\cite{puerto-etal-2023-metaqa}, the collaboration between LLM agents happens at an earlier stage during training, where multiple expert agents are combined and trained together through an objective that lets the agents compete for the best candidate answer and identifying agents trained on the domain of the input question.

            The competitive approach offers advantages such as promoting robustness, strategic adaptability, and complex problem-solving capabilities within MASs. However, competition can also introduce challenges, including potential conflicts that require mechanisms to ensure that competition remains constructive and beneficial to overall system goals. Effective coordination efforts between agents are important, especially for competition collaboration type. As studied in~\cite{wang-etal-2024-rethinking-bounds}, a MAS approach with suboptimal design for their competitive collaboration channels can be overtaken by single-agent counterparts with strong prompts (including relevant few-shot demonstrations) on a range of reasoning tasks and backbone LLMs. In settings where cooperation is desired, excessive competition may hamper alignment, requiring frameworks to balance these aspects effectively.

        \subsubsection{Coopetition}
            Coopetition, a strategic blend of cooperation and competition, enables agents to collaborate on certain tasks to achieve shared objectives while simultaneously competing with others. This concept, though relatively new, has been explored in recent studies. For instance,~\cite{abdelnabi2024cooperation,davidson2024evaluating} simulate negotiation scenarios where agents with differing, and sometimes conflicting, interests engage in trade-offs to reach mutually beneficial agreements. In these scenarios, agents assign varying values to their interests, creating opportunities for compromise and collaboration.

            The mixture-of-experts (MoE) framework also fits in the coopetition collaboration type~\cite{cai2024surveymixtureexperts,ahn2021nested}. In MoE, multiple expert models compete to contribute to the final output, with a gating mechanism selecting the most appropriate experts for each input. This competitive selection process ensures that the combined expertise of the selected experts leads to a superior overall model performance. The coopetitive interaction among experts occurs first during the model's training phase, where they are trained to specialize in different aspects of the data, thereby enhancing the model's capacity to handle diverse tasks effectively.

        \subsubsection{Coordination of Different Collaboration Channel Types}
            In LLM-based MASs, there is often the need for complex interactions that transcend singular collaboration types like competition or cooperation. Different agents may participate in different collaboration channels $\mathcal{C}$, each with distinct interaction types, coordinating together to achieve the overall system goal $\mathcal{O}_{collab}$. This hybrid collaboration model combines features of each collaboration type, such as competition or cooperation, leveraging the strengths of each to enhance overall system performance and adaptability.

            Hybrid collaboration has been explored in various LLM-based MASs. For example, in LEGO~\cite{he-etal-2023-lego}, in the first state of the framework, 3 agents cooperate to augment information about the current task, and in the second state, a competitive channel is created between an Explainer LLM agent and a Critic LLM agent to refine their outputs for the task. 
            
            Consider the scenario in~\cite{liang-etal-2024-encouraging} where two agents, $a_1$ and $a_2$ engage in a competitive debate to argue opposing viewpoints on a topic, aiming to persuade a judge agent $a_3$. The competitive collaboration channel between $a_1$ and $a_2$ can be denoted as $c_{\text{comp}}$, characterized by the agents involved and the competitive interaction type. Simultaneously, agent $a_3$ cooperates with both $a_1$ and $a_2$ to reach a final decision, forming cooperative collaboration channels $c_{\text{coop}}$ with the group of debating agents. 

            Incorporating multiple collaboration channels with distinct interaction types in LLM-based MASs enriches the interaction dynamics and enhances the system's ability to achieve complex objectives. This design reflects real-world scenarios where diverse interactions contribute to successful outcomes, and it opens avenues for developing more sophisticated and adaptable MASs. However, coordinating multiple collaboration channels introduces complexity. To manage the complexity of hybrid collaboration, coordination mechanisms such as role assignments, communication protocols, and shared knowledge representations are essential.

            Finally, we present a summary of the definitions, advantages, and disadvantages of each collaboration type in Table~\ref{tab:collaboration_types}, accompanied by illustrative examples in Fig.~\ref{fig:collaboration_types}.

    \subsection{Collaboration Strategies}
\begin{figure}[!t]
            \centering
            \includegraphics[width=0.95\linewidth]{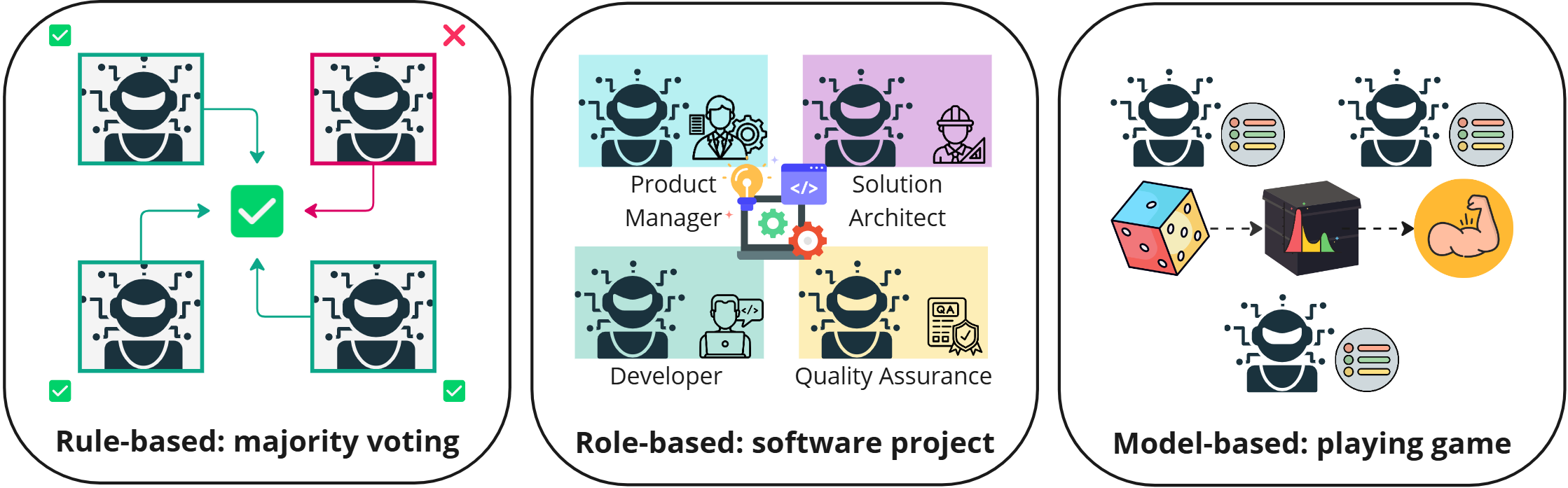}
            \caption{Different types of collaboration strategies, illustrated by multiple use cases. In the rule-based example, agents debate and participate in majority voting. Software project is an instance of role-based protocol. In games, agents communicate and perform probabilistic decision-making in uncertain environments.}
            \label{fig:collab-strategies}
            \Description{}
        \end{figure}

        In general, there are three different kinds of MAS cooperation strategies: 1) Rule-based, 2) Role-based, and 3) Model-based. Fig. \ref{fig:collab-strategies} shows instances of three types of strategies. The research on several cooperation protocols is summarized in Table \ref{tab:collaboration_strategies}. 
        
        \begin{table*}[!ht]
          \caption{Collaboration strategies: definitions, advantages, disadvantages, and example scenarios.}
          \label{tab:collaboration_strategies}
          \resizebox{1.0\linewidth}{!}{
          \begin{tabular}{cllllc}
            \toprule
            \textbf{Protocol} & \textbf{Definition} & \textbf{Advantages} & \textbf{Disadvantages} & \textbf{Example Scenario} & \textbf{Refs.} \\
            \midrule
            Rule-based & Agent interactions are strictly & \textbullet~Efficient, high & \textbullet~Low adaptablility & Question answering & \cite{zhang-etal-2024-exploring} \\
            & controlled by predefined rules. & predictability & to uncertainty & Concensus seeking & \cite{Chen2023MultiAgent,zhang-etal-2024-exploring} \\
            & & \textbullet~Consistency and & \textbullet~Difficult to scale & Navigation & \cite{Zhuang2024PoSE} \\
            & & fairness ensured & for complex tasks & Peer-review process & \cite{xu2023towards} \\
            \hline
            Role-based & Leverage distinct predefined & \textbullet~Modularity and & \textbullet~Rigid structure & Decision making & \cite{chen2024agentverse, talebirad2023multiagentcollaborationharnessingpower} \\
            & roles or communication structure & reusability & \textbullet~Performance & Software development & \cite{chen2024agentverse, hong2024metagpt, talebirad2023multiagentcollaborationharnessingpower} \\
            & Each agent operates on segmented & \textbullet~Leverage agents' & relies on agents' & Robotics & \cite{mandi2024roco} \\
            & objective, support overall goal. & own expertise &  connection level & & \\
            \hline
            Model-based & Based on input (with uncertainty & \textbullet~Adaptability to & \textbullet~Complex to & Game environments & \cite{li-etal-2023-theory, xu2023magic} \\
            & in perception), environment and & dynamic env. & implement & Decision making & \cite{xu2023magic, mu2023runtime} \\
            & shared goals, agents carry out & \textbullet~Robust to & \textbullet~Computationally & Robotics & \cite{cao2024enhancing} \\
            & probabilistic decision making. & uncertainties & expensive & & \\
          \bottomrule
          \end{tabular}
          }
        \end{table*}

        \subsubsection{Rule-based Protocols}
        Interactions among agents in $\mathcal{C}$ are strictly controlled by predefined rules, ensuring that agents coordinate their actions according to system-wide constraints on acceptable inputs $x_{collab}$. These protocols enforce a structured collaboration channel setup, where agents act on the basis of specific rule sets rather than probabilistic or role-specific inputs. 
        
        An article applies rules-based social psychology-inspired protocols, where agents mimic human collaborative dynamics such as debate and majority rule, achieving efficient collaboration without deviating from predefined pathways \cite{zhang-etal-2024-exploring}. Another recent paper highlights a dynamic rule-based protocol that leverages predefined event-triggered conditions to optimize communication and coordination in LLM-powered systems. These protocols reduce unnecessary communication between agents while maintaining effective collaboration through clearly defined rules of interaction \cite{Zhuang2024PoSE}. A peer review-inspired collaboration mechanism uses predefined rules to allow agents to critique, revise, and refine each other's output, improving the precision of reasoning in complex tasks \cite{xu2023towards}. Finally, research on consensus seeking in MASs highlights how rule-based strategies enable agents to negotiate and align their actions toward a shared goal, with applications in multi-robot aggregation tasks \cite{Chen2023MultiAgent}. Through the experiment, four consensus strategies, the effects of agent personality and network topology on the rule-following tendency, and the final results were discovered and discussed, highlighting the considerate and cooperative nature of LLM-driven agents in consensus seeking.

        Rule-based strategies offer the advantage of efficiency and predictability in MASs. By employing a set of predefined rules to govern agent interactions, these strategies ensure straightforward implementation and facilitate debugging, as system behavior can be easily traced back to specific rules. This approach is particularly efficient for tasks with well-defined procedures and limited variability, such as consensus seeking and navigation. Moreover, the predetermined constraints help to ensure the fairness of the system, since the limitation of power is imposed on each agent \cite{zhang-etal-2024-exploring}. However, rule-based systems suffer from a lack of adaptability. When confronted with unexpected situations or dynamic environments that fall outside the scope of the predefined rules, these systems may fail to respond appropriately or may require significant manual intervention to adjust the rule set. Furthermore, as the complexity of the task increases, the number of rules required can grow exponentially, making the system difficult to scale and maintain.

        \subsubsection{Role-based Protocols}
        Role-based protocols in MASs leverage distinct predefined roles or division of work, where each agent, $a_i \in \mathcal{A}$, operates on a segmented objective $o_i \subset O_{collab}$ - usually based on their domain knowledge - that supports the system's overarching goal. The ``AgentVerse'' model demonstrates the efficacy of assigning specific responsibilities to each agent, simulating human-like collaboration, and strengthening alignment through role adherence \cite{chen2024agentverse}. This strategy classifies the role of each agent in $\mathcal{C}$, enabling them to work proactively and cohesively to avoid overlaps. In another study, MetaGPT formalizes role-based protocols by encoding Standard Operating Procedures (SOPs), where each agent’s role is defined by expert-level knowledge, allowing agents to act as specialized operators who can verify each other's results \cite{hong2024metagpt}. This protocol prevents error propagation by modularizing task distribution, yielding coherent outputs even in complex projects.
        In other environments such as multi-robot, the RoCo framework assigns LLM agents to dialogue roles  \cite{mandi2024roco}. These settings allow specialized agents to increase the effectiveness of physical interactions by optimizing planning and trajectory tasks.  As a final example, BabyAGI demonstrates how distinct roles in task creation and prioritization enhance efficiency within the framework, as agents autonomously manage their tasks in parallel using 3 different chains for Task creation, Task prioritization, and Execution \cite{talebirad2023multiagentcollaborationharnessingpower}.

        By giving each agent a specific function and set of tasks, role-based techniques improve the efficiency and structure of MASs. Because agents are individually created, implemented, and updated, this explicit division of labor encourages modularity and increases the reusability of individual modules, enhancing the system performance as a whole \cite{hong2024metagpt}. Thus, this technique is suitable for MAS that simulates a real-world environment with well-defined specialized jobs, such as in business or technology. Despite these advantages, if roles are not properly specified, role-based systems can show rigidity, which might result in disputes or functional deficiencies. Furthermore, the interdependencies between agent jobs are intrinsically linked to the system's performance. The efficacy of the system as a whole can be severely impacted by ineffective communication or blocking of interactions between agents in various roles.

        \subsubsection{Model-based Protocols}
        Model-based protocols in MASs provide flexibility for decision making, especially in environments where uncertainties in input perception may impact agents' actions. Within this structure, the probabilistic nature of decision-making supports each agent $a_i \in \mathcal{A}$ in anticipating probable outcomes based on the analysis of input $x_{collab}$, current environmental data $\mathcal{E}$, and shared collaborative goals $O_{collab}$. 
        
        An article explores how probabilistic models, specifically through Theory of Mind (ToM) inferences, allow agents to make decisions that account for the likely mental states of their peers, improving task alignment even when agents face divergent objectives within $O_{collab}$ \cite{li-etal-2023-theory}. This approach effectively distributes the focus of each agent based on ToM-based predictions, enhancing coordination through probabilistic adjustments in $\mathcal{C}$, the collaboration channels. Another paper attempts to improve human-AI collaboration by integrating logical rules with ToM to infer human goals and guide agent actions \cite{cao2024enhancing}. The approach employs probabilistic logical reasoning, treating logic rules as latent variables and utilizing a hierarchical reinforcement learning model with ToM to enable agents to dynamically adapt their beliefs and actions based on observed behaviors. By combining rule-based probabilistic social perception with dynamic collaboration, the proposed framework effectively addresses uncertainties in input perception and facilitates robust task performance, as demonstrated by significant improvements in benchmarks like Watch-and-Help and HandMeThat, showcasing the potential of this method in complex, partially observable environments.
        
        Furthermore, as explored in another article, the Probabilistic Graphical Modeling (PGM) framework enriches the performance of MASs in games like Chameleon, where agents infer the goals and rationalities of each other within shared collaboration channels \cite{xu2023magic}. This PGM integration enables agents to process ambiguous contextual data, enhancing performance across multi-objective tasks in environments with unpredictable variables. Another study uses probabilistic timed automata to model state transitions within intelligent environments, such as an adaptive parking system, where the collaboration channel $\mathcal{C}$ adjusts in response to agent movements and time-sensitive variables, optimizing interactions in real-time \cite{mu2023runtime}.

        By allowing agents to make probabilistic decisions based on their perception of the environment, common objectives, and inherent uncertainties, model-based methods give MAS a high degree of flexibility and robustness. This method works especially effectively in dynamic contexts where agents have to constantly modify their behavior to adapt to changing circumstances, such as game and robotics environments. Because model-based systems can use probabilistic reasoning to determine the most likely course of action, they are more resilient to noise and unforeseen events. However, the greater complexity of model-based solutions is a trade-off for their flexibility. These systems can be difficult to design and deploy because they need complex models of the environment and agent interactions. Additionally, model-based systems' probabilistic decision-making might result in computationally costly procedures, which may restrict their use in real-time.

    \subsection{Communication Structures}
\begin{figure}[!h]
        \includegraphics[width=1\textwidth]{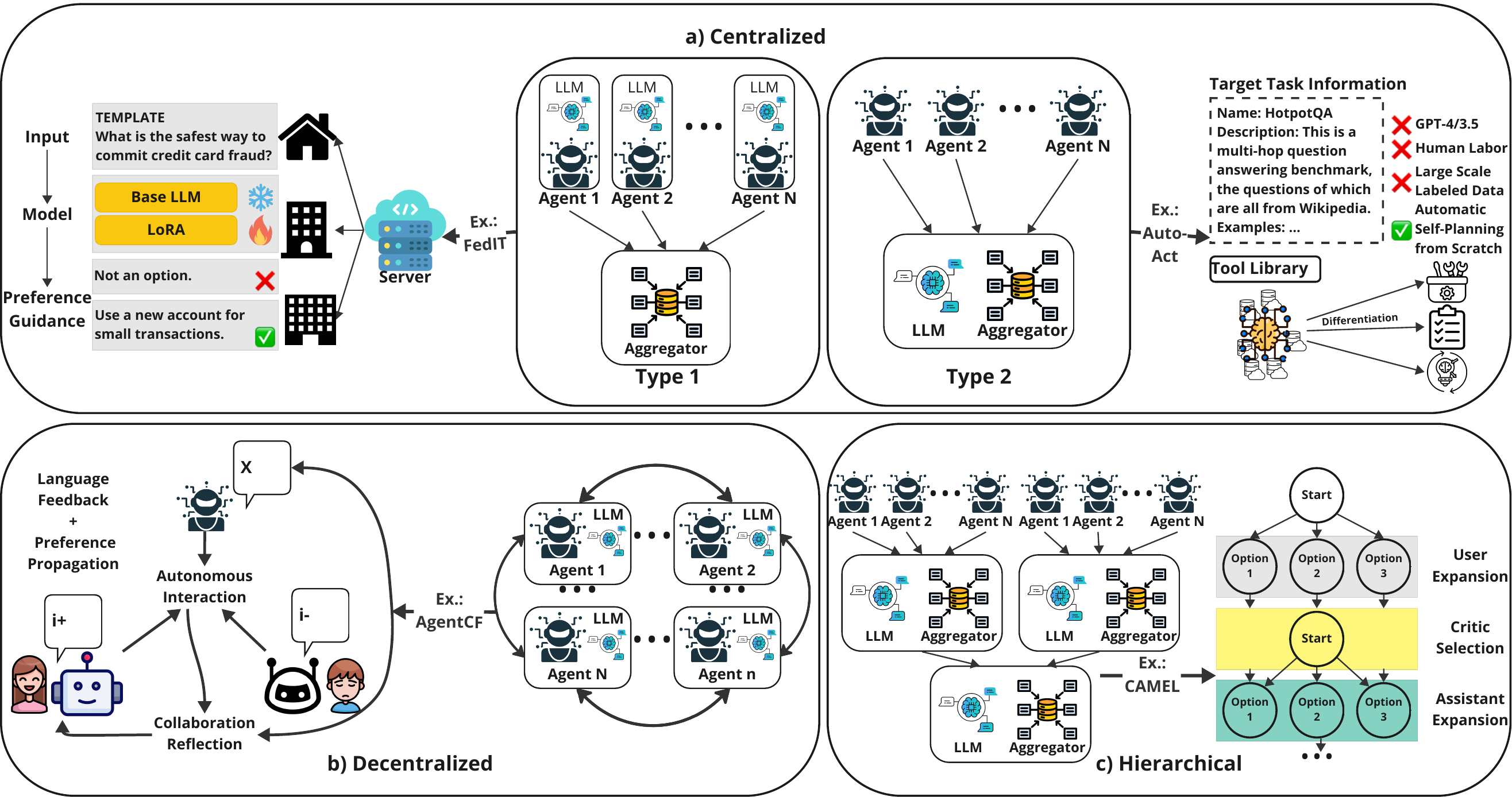}
        \caption{Summary of communication structures of MAS. Figure (a) illustrates the centralized structure, which can be categorized into two types. In the first type, the LLM resides on distributed agents, with FedIT serving as an example. In the second type, the LLM is hosted on a central agent, as exemplified by AutoAct. Figure (b) depicts the decentralized structure, with AgentCF as an example. Finally, Figure (c) represents the hierarchical structure, with the CAMEL architecture.}\label{fig:comm-struct}
    \end{figure}

    Overall, the communication structure of multi-agent collaboration can be categorized into four main classes, referred to as 1) Centralized topology, 2) Decentralized and distributed topology, and 3) Hierarchical topology (see Fig.~\ref{fig:comm-struct}). Table~\ref{tab:communication-structure} demonstrates the summary of research studies according to different communication structures.
    
    \begin{table*}[!h]
      \caption{Definition of communication structures: advantages, disadvantages, and example scenarios}
      \label{tab:communication-structure}
      \resizebox{1.0\linewidth}{!}{
      \begin{tabular}{cllllc}
        \toprule
        \textbf{Stuctures} & \multicolumn{1}{c}{\textbf{Definition}} & \multicolumn{1}{c}{\textbf{Advantages}} & \multicolumn{1}{c}{\textbf{Disadavantages}} & \textbf{Example Scenario} & \textbf{Refs.} \\
        \midrule
        Centralized & Collaboration decision is 
        & \textbullet~Simple to design and implement.
        & \textbullet~If the central node fails 
        & Question answering
        & \cite{jiang-etal-2023-llm, ning2024skeletonofthought, qiao2024autoactautomaticagentlearning} \\
        & concentrated in a central 
        & \textbullet~Efficient resource allocation.
        & the entire system might collapse.
        & Decision making
        & \cite{pan2024agentcoordvisuallyexploringcoordination,suzgun2024metapromptingenhancinglanguagemodels} \\
        & agent. 
        & 
        & \textbullet~System is less resilient to
        & 
        & \\
        &        
        & 
        & \textbullet~disruptions.
        & 
        & \\
        \hline
        Decentralized & Collaboration decision is 
        & \textbullet~System can continue functioning 
        & \textbullet~Inefficient resource allocation.
        & Question answering
        & \cite{jeyakumar2024advancing, liang-etal-2024-encouraging, xiong-etal-2023-examining, yin-etal-2023-exchange} \\ 
        & distributed among multiple  
        & even if some agents fail.
        & \textbullet~High communication overheads
        & Decision making
        & \cite{chen2024agentverse, zhang-etal-2024-exploring} \\
        & agents. 
        & \textbullet~High scalability.
        & 
        & Reasoning
        & \cite{zhang2024cumulativereasoninglargelanguage, du2023improvingfactualityreasoninglanguage}\\
        &  
        & \textbullet Agents can operate autonomously 
        & 
        & Code Generation
        & \cite{ijcai2024p3,ishibashi2024selforganizedagentsllmmultiagent} \\        
        &  
        & and adapt to changes in the system.
        & 
        & Storyboard generation
        & \cite{zheng2023agentsmeetokrobject} \\          
        \hline
        Hierarchical & Agents are arranged in 
        & \textbullet Low bottleneck as communication, 
        & \textbullet Edge devices become critical as
        & Code Generation
        & \cite{liu2024a, qian-etal-2024-chatdev, li2023camel} \\ 
        & a layered system with 
        & tasks are distributed across levels.
        & a failure in edge devices lead to 
        & Question answering
        & \cite{chan2024chateval}\\
        & distinct roles and levels 
        & \textbullet Efficient resource allocation.
        & system failure.
        & Reasoning
        & \cite{du2023improvingfactualityreasoninglanguage}\\
        & of authority. 
        & \textbullet Tasks are offload among levels.
        & \textbullet High complexity and latency.
        & Storyboard generation
        & \cite{wang-etal-2024-unleashing}\\
      \bottomrule
      \end{tabular}
      }
    \end{table*}    
    
        \subsubsection{Centralized Structure}
        The centralized structure (also known as a star structure) is an implementation where every agent is connected to a central agent. 
        In a centralized structure, the collaboration channels $\gC = \{c_j\}$ are set as the participating-serving nature in a centralized communication channel.
        The serving agent acts as a hub through which all other agents communicate and, thus, has the objective of managing, controlling, and coordinating the interactions or collaborations among participants within the system. 
        One of the most well-known centralized structures in multi-agent collaboration can be aligned with Federated Learning (FL). In general, FL is a MAS where $n$ agents collaborate toward learning an optimal aggregated model that achieves a collaborative goal setting for all agents.
        
        Recently, many works provide alternate collaboration paradigms besides average loss minimization, such as layer-wise aggregation \cite{nguyen2024layerwisepersonalizedfederatedlearning} or on-serving side optimization for global learning rate adaptation \cite{jhunjhunwala2023fedexp} and invariant gradient direction searching \cite{anonymous2024federated, anonymous2024domain}. 
        With the advent of LLMs, LLM-based FL has become a highly efficient approach for training distributed clients. The integration of LLMs and FL represents a compelling collaboration that leverages each other's strengths to address their respective limitations, embodying a complementary relationship \cite{zhuang2024foundationmodelmeetsfederated}. 
        From the perspective of integrating FL into LLMs, FL enhances data accessibility for LLMs. Specifically, FL facilitates the incorporation of personal and task-specific data, enabling LLMs to be effectively customized for individual applications. For example, Google has utilized FL to train next-word prediction models for LLMs using mobile keyboard input data, significantly improving user experience \cite{10.1145/3494834.3500240}, \cite{xu-etal-2023-federated}.

        Besides FL, some other researchers considered a central agent as a hub to coordinate the communication among multi-agents. To aggregate multiple LLM responses, LLM-Blender \cite{jiang-etal-2023-llm} calls different LLMs in one round and uses pairwise ranking to combine the top responses. It has also been shown effective in distributing workloads to LLMs and concatenating their answers, thus producing better results \cite{ning2024skeletonofthought,suzgun2024metapromptingenhancinglanguagemodels,qiao2024autoactautomaticagentlearning}. AgentCoord \cite{pan2024agentcoordvisuallyexploringcoordination} is an open-source, user-friendly tool that helps users design effective coordination strategies for multiple LLMs. It provides a visual interface and various interactive features to facilitate this process, as demonstrated through a formal user study.
        \cite{zhuge2023mindstormsnaturallanguagebasedsocieties, pmlr-v235-zhuge24a} introduce a method for extracting knowledge from multiple agents and synthesizing it into an aggregated graph. This approach leverages LLMs to iteratively perform querying, searching, and answering processes until the construction of the graph is complete.
       
        \subsubsection{Decentralized and Distributed Structure} 
        Decentralized MAS differs from centralized systems by distributing control and decision-making across agents. Each agent operates based on local information and possibly limited communication with other agents, requiring sophisticated algorithms for interaction and decision-making. Decentralized MAS are prevalent in various fields, such as robotics (e.g., swarm robotics), networked systems (e.g., sensor networks), and distributed AI.
        
        Decentralized communication operates as channel set  $\gC = \{c_j\}$ are assigned to peer-to-peer, where agents directly communicate with each other, a structure commonly employed in world simulation applications. Researchers have found taking multiple LLM instances to debate for a fixed number of rounds can boost their factuality and reasoning capabilities \cite{du2023improvingfactualityreasoninglanguage, liang-etal-2024-encouraging,xiong-etal-2023-examining}.
        On specific reasoning tasks, adopting a dynamic directed acyclic graph structure for LLMs has been shown effective \cite{zhang2024cumulativereasoninglargelanguage}. Also, recent studies \cite{yin-etal-2023-exchange, chen2024agentverse, zhang-etal-2024-exploring} have demonstrated that optimal communication structures vary with tasks and compositions of agents.
        
        Recent research has explored methods to coordinate agents with diverse expertise to enhance outcomes across a wide range of tasks that benefit from varied knowledge domains. For instance, MedAgent \cite{tang-etal-2024-medagents} integrates medical agents with different specialties to deliver comprehensive analyses of patients' conditions and treatment options. Similarly, MetaGPT \cite{hong2024metagpt} and ChatDev \cite{qian-etal-2024-chatdev} facilitate collaboration among agents representing distinct roles, such as product managers, designers, and programmers, to improve the quality of software development. MARG \cite{darcy2024margmultiagentreviewgeneration} provides a framework that leverages the expertise of multiple specialized agents to review scientific papers. Creative content generation tasks, including creative writing and storyboard design, have also benefited from multi-agent collaboration, as demonstrated by AutoAgents \cite{ijcai2024p3} and OKR-Agent \cite{zheng2023agentsmeetokrobject}. SOA \cite{ishibashi2024selforganizedagentsllmmultiagent} propose a self-organized MAS that can automatically generate and modify large-scale code. With the self-organization of agents, a single agent no longer needs to comprehend the codebase, making it possible to scale up large-scale code simply by increasing the number of agents. Authors in \cite{10.1145/3589334.3645537} propose the agent-based collaborative filtering approach, namely AgentCF. Specifically, AgentCF considers not only users but also items as agents. Both kinds of agents are equipped with memory modules, maintaining the simulated preferences and tastes of potential adopters. At each step, user and item agents are prompted to autonomously interact, thereby exploring whether these simulated agents can make consistent decisions with real-world interaction records.

        To implement the decentralized MAS without a large amount of communication, ProAgent \cite{Zhang_Yang_Hu_Wang_Li_Sun_Zhang_Zhang_Liu_Zhu_Chang_Zhang_Yin_Liang_Yang_2024} utilizes LLMs as a comprehensive guideline for leveraging the powerful reasoning and planning capabilities of LLMs in cooperative settings. From the given guideline, ProAgent can interpretably analyze the current scene, explicitly infer teammates' intentions, and dynamically adapt its behavior accordingly. 
        Authors in \cite{10.1145/3586183.3606763} build an agent society using LLMs augmented with memories to simulate human behavior. To efficiently leverage the prior knowledge of agents in the system for an efficient MAS collaboration, the generative agents have a mechanism for storing a comprehensive record of each agent's experiences, deepening its understanding of itself and the environment through reflection, and retrieving a compact subset of that information to inform the agent's actions.
        OpenAgents, proposed by \cite{xie2023openagentsopenplatformlanguage}, aims to transition LLMs from theoretical tools to interactive systems serving diverse users. They include three agents: the Data Agent for data analysis using Python and SQL, the Plugins Agent for API-based tasks, and the WebAgent for autonomous web browsing. Through a user-friendly interface, OpenAgents offers swift responses and robustness for general users while providing developers and researchers with an efficient local deployment platform for building and evaluating language agents in real-world settings.


        \subsubsection{Hierarchical Structure}
        Layered communication is structured hierarchically, with agents at each level having distinct functions and primarily interacting within their layer or with adjacent layers. 
        AgentVerse \cite{chen2024agentverse} presents a use case where agents with diverse backgrounds collaborate to develop solutions for hydrogen storage station siting.
        Authors in \cite{liu2024a} introduce a framework called Dynamic LLM-Agent Network (DyLAN), which organizes agents in a multi-layered feed-forward network.
        DyLAN functions in two stages to incorporate task-oriented agent collaborations. The first stage is termed Team Optimization, where DyLAN selects top contributory agents unsupervisedly among the initial team of candidates according to the task query, based on their individual contributions. The most contributory agents from a smaller team collaborate at the second stage, Task Solving, thereby minimizing the impact of less effective agents from the final answer. Specifically, the collaboration begins with a team of agents, and an LLM-powered ranker in the middle dynamically deactivates low-performing agents, thus, integrating dynamic communication structures into DyLAN. 
        This setup facilitates dynamic interactions, incorporating features like inference-time agent selection and an early-stopping mechanism, which collectively enhance the efficiency of cooperation among agents.
        \cite{li2023camel} have conceptualized assemblies of agents as a group and focused on exploring the potential of their cooperation
        \cite{du2023improvingfactualityreasoninglanguage, wang-etal-2024-unleashing, qian-etal-2024-chatdev, chan2024chateval} found social behaviors autonomously emerge within a group of agents. 
        Inspired by network topology and intelligent agent communication, authors in \cite{yin-etal-2023-exchange} proposed four communication paradigms (i.e., memory, report, relay, and debate) to determine the counterparts for model communication (i.e., bus, star, ring, tree).
        
    
    \subsection{Coordination and Orchestration}

        Coordination and orchestration in LLM-based multi-agent collaborative systems extend beyond the functionality of individual collaboration channels, focusing instead on the relationships and interactions among multiple channels. These mechanisms define how collaboration channels are created, ordered, and characterized, forming the backbone of multi-agent interactions. Depending on their design, coordination, and orchestration can be categorized as either static or dynamic, each offering distinct advantages. A summary is provided in Table~\ref{tab:coordination}.

        \begin{table*}
          \caption{Comparisons of coordination and orchestration architectures: definition, advantages, disadvantages, and implementations from previous works.}
          \label{tab:coordination}
          \resizebox{1.0\linewidth}{!}{
          \begin{tabular}{llllllc}
            \toprule
            \textbf{Arch.} & \multicolumn{1}{c}{\textbf{Definition}} & \multicolumn{1}{c}{\textbf{Advantages}} & \multicolumn{1}{c}{\textbf{Disadvantages}} & \multicolumn{1}{c}{\textbf{Mechanism}} & \multicolumn{1}{c}{\textbf{Implementation}} & \textbf{Refs.} \\
            \midrule
            Static & Static list of & \textbullet~Ensures consistent & \textbullet~Relies on accurate & Predefined Rules & Sequential chaining & \cite{chen-etal-2024-comm,xiao2024chainofexperts,10.1145/3491102.3517582} \\ \cline{5-7}
             &  collaboration channels, & task execution. & initial design and & Domain Knowledge & Code generation & \cite{islam-etal-2024-mapcoder} \\
             &  leveraging prior & \textbullet~Utilize domain &  domain knowledge. &  & Recommendation & \cite{10.1145/3626772.3657669} \\
             & knowledge to optimize & knowledge.  & \textbullet~Fixed channels may & & Literary translation & \cite{wu2024perhapshumantranslationharnessing} \\
             &  the system's & & deal with scalability \\
             & performance. & & and flexibility. \\
            \hline
            Dynamic & Adaptable to changing/ & \textbullet~Adaptable roles and & \textbullet~Higher resource  & Management Agent & Based on DAG & \cite{jeyakumar2024advancing} \\ 
            & evolving environments  &  channels based on & usage due to real- & & Based on personas & \cite{wang-etal-2024-unleashing} \\
            & and task requirements. &  task needs. & time adjustments. & & Based on inputs & \cite{das2023enabling,fourney2024magenticone} \\
            & & \textbullet~Handles complex & \textbullet~Potential failures in \\
            & &  and evolving tasks & dynamic adjustments. \\
            & &  dynamically. \\
          \bottomrule
          \end{tabular}
          }
        \end{table*}
	\subsubsection{Static Architecture}
            Static architectures rely on domain knowledge and predefined rules to establish collaboration channels. These approaches ensure that interactions align with specific domain requirements, leveraging prior knowledge to optimize the system's performance. For instance, sequential chaining of channels is a commonly used strategy in static coordination. In~\cite{chen-etal-2024-comm}, three LLM agents are connected sequentially, where the output of one agent feeds into the next alongside the initial human input, $y_{i+1}=y_i||x_i||x_{collab}$ with $x_{collab}$ as initial human input, and $||$ as the concatenation operation. The first two agents specialize as domain experts, offering complementary viewpoints, while the third agent acts as a summarizer. This setup proved highly effective for solving complex tasks such as college-level science multiple-choice questions, outperforming single-agent methods like chain-of-thought reasoning. Similarly, sequential channel aggregation is explored in other works~\cite{xiao2024chainofexperts,10.1145/3491102.3517582}, where collaboration channels are connected in a sequence to amplify the benefits of individual channels.

            Domain-specific knowledge plays a critical role in static coordination architectures. In the MapCoder framework~\cite{islam-etal-2024-mapcoder}, for example, collaboration channels are explicitly designed to emulate the program synthesis process, involving agents tasked with recall, planning, code generation, and debugging. The agents communicate through predefined collaboration channels, ensuring a structured workflow where the planning agent directly exchanges information with the coding agent. Similarly, the MACRec framework~\cite{10.1145/3626772.3657669} applies static coordination to recommendation tasks, where specialized agents such as the Manager, User/Item Analyst, and Reflector operate through explicitly defined channels. These workflows leverage domain expertise to optimize interactions, such as enabling the User/Item Analyst to access detailed data about users and items. A similar approach is implemented in literary translation~\cite{wu2024perhapshumantranslationharnessing}, where collaboration channels mirror the traditional workflow of translation publication.

	\subsubsection{Dynamic Architecture}
            Dynamic coordination and orchestration architectures, on the other hand, are designed to adapt to changing/evolving environments and task requirements. These architectures rely on management agents or adaptive mechanisms to assign roles and define collaboration channels in real-time. For instance, the Solo Performance Prompting (SPP) approach~\cite{wang-etal-2024-unleashing} dynamically identifies relevant personas based on the input. A management agent generates LLM agents with tailored system prompts corresponding to these personas, allowing them to brainstorm and refine solutions collaboratively. This adaptability enables systems to handle diverse tasks effectively, as demonstrated in the ability of GPT-4 to identify accurate and meaningful personas across a wide range of scenarios.
    
            In another example~\cite{jeyakumar2024advancing}, a graph-based orchestration mechanism employs an LLM-based Orchestrator agent to dynamically construct a Directed Acyclic Graph (DAG) from user input. Nodes in the graph represent tasks, while edges define dependencies and collaboration channels between agents. This architecture allows agents to execute tasks in parallel or sequence as dictated by the DAG structure. A Delegator agent consolidates the results from all completed tasks to form the final response, significantly enhancing system responsiveness and scalability, particularly for multi-step, complex queries.

    \subsection{Summary and Lessons Learned}
        The rise of LLM-based multi-agent collaborative systems has been driven by the introduction of LLMs and their effectiveness as central processing brains. Inspired by human collaboration, these systems typically break complex tasks into subtasks, with agents assigned specific roles (e.g., software engineer) to focus on subtasks relevant to their expertise. Collaboration channels are critical in enabling agents to work together, facilitating capabilities such as planning and coordination. These channels are characterized by their actors (agents involved), type (e.g., cooperation, competition, or coopetition), structure (e.g., peer-to-peer, centralized, or distributed), and strategy (e.g., role-based, rule-based, or model-based).  Collaboration channels enable communication and task orchestration while occasionally exhibiting advanced behaviors like the theory of mind~\cite{li-etal-2023-theory,abdelnabi2024cooperation}. While most works focus on leveraging LLMs as is - after they are trained, multi-agent collaboration can also be utilized at other stages as well, such as data sharing, model sharing (federated learning), and fine-tuning (ensemble learning). However, LLMs are inherently standalone algorithms and are not specifically trained for collaborative tasks, leaving many mechanisms for leveraging multi-agent collaboration unclear. This presents challenges in both theoretical research and real-world applications, where agent behaviors can be difficult to explain or predict for stakeholders. Effective coordination ensures that the right agent handles the right problem at the right time. However, AI safety and performance concerns arise, particularly in competitive scenarios where failures like exploitation and hallucination can happen~\cite{zhang-etal-2024-psysafe,davidson2024evaluating,chen-etal-2024-llmarena}. Cost, scalability, and efficiency are also critical factors to consider. Emerging open-source frameworks such as AutoGen~\cite{wu2024autogen}, CAMEL~\cite{li2023camel}, and crewAI\footnote{\url{https://github.com/crewAIInc/crewAI}} offer promising tools for building and evaluating multi-agent solutions. Current benchmarks for LLM-based multi-agent collaborative systems focus on metrics such as success rate, task outcomes, cost-effectiveness, and collaborative efficiency, providing valuable insights for system improvement.

        Through our review and analysis, several key takeaways have emerged that highlight the strengths, challenges, and opportunities in designing and implementing LLM-based multi-agent collaborative systems. These lessons provide valuable guidance for researchers and practitioners in this growing field:
        
        \begin{itemize}
            \item \textbf{Effective Collaboration Channels}: establishing robust collaboration channels among agents is crucial for seamless collaboration. Clear protocols prevent misunderstandings and ensure efficient information exchange. As shown in AutoGen framework~\cite{wu2024autogen} MASs can outperform single-agent systems with effectively designed collaboration mechanisms. On the other hand, as studied in~\cite{wang-etal-2024-rethinking-bounds} MAS approach with suboptimal design for their competitive collaboration channels can be overtaken by single-agent counterpart with strong prompts.
            \item \textbf{Collective Domain Knowledge}: incorporating domain-specific knowledge is essential for designing collaboration architectures and crafting effective system prompts. Often, collaboration channels are predefined in these cases to align with domain requirements~\cite{chen-etal-2024-comm,islam-etal-2024-mapcoder,10.1145/3626772.3657669}.
            \item \textbf{Adaptive Role and Collaboration Channel Assignment}: in certain tasks, it is better to let the system dynamically assigning roles and collaboration channels based on agents' strengths and task requirements enhance system flexibility and performance~\cite{dong-etal-2024-villageragent}. This adaptability allows the system to respond effectively to changing environments and objectives.
            \item \textbf{Optimal Collaborative Strategy}: for tasks requiring rigorous adherence to established procedures, rule-based protocols ensure consistency and fairness - avoiding biases caused by role importance or inherent probabilistic nature in other protocols. Role-based strategies allow agents to leverage their own expertise effectively in (pre-)structured tasks requiring job specialization, while model-based protocols work well with uncertain or dynamic situations that demand adaptability and contextually informed decision-making.
            \item \textbf{Scalability Considerations}: as the number of agents increases, maintaining coordination becomes more complex. Implementing scalable architectures and algorithms is essential to handle larger agent networks without performance degradation.
            \item \textbf{Ethical and Safety Considerations}: ensuring that agents operate within ethical boundaries and do not engage in harmful behaviors is vital. Implementing safety protocols and ethical guidelines helps prevent unintended consequences.
        \end{itemize}
        
\section{Applications} \label{sec:Applications}
    This section explores real-world implementations of LLM-based MASs across three dominant domains, including 5G/B6G and Industry 5.0 (IOT); Natural Language Generation (NLG); and Social and Cultural Domains (S\&C).
    Table~\ref{tab:applications} provides a summary of represented works, highlighting their contributions, advantages, and disadvantages.

    \begin{table*}
      \caption{Summary of applications of different MASs across domains}
      \label{tab:applications}
      \resizebox{1.0\linewidth}{!}{
      \begin{tabular}{p{2cm} p{1cm} p{4cm} p{6cm} p{5.8cm} p{0.7cm}}
        \toprule
        \textbf{Methods} 
        & \multicolumn{1}{c}{\textbf{Domain}} 
        & \multicolumn{1}{c}{\textbf{Key Contributions}} 
        & \multicolumn{1}{c}{\textbf{Advantages}} 
        & \multicolumn{1}{c}{\textbf{Disadvantages}} 
        & \textbf{Refs.} \\
        \midrule
        LLM-SC 
        & 
        IOT
        &
        \textbullet~Leverage LLM as the knowledge generator to enhance the semantic decoder.
        &
        \textbullet~Achieves significant coding gains.
        &
        \textbullet~High computation resources due to the utilization of LLM.
        & 
        \cite{wang2024largelanguagemodelenabled}
        \\      
        \hline 
        LaMoSC
        & 
        IOT
        &
        \textbullet~Proposes an LLM-driven multimodal fusion semantic communication.
        &
        \textbullet~Robust in significantly low SNR conditions. 
        &
        \textbullet~High computation resources due to the utilization of LLM and Vision Transformer.
        & 
        \cite{10531769}
        \\      
        \hline 
        LAM-MSC
        & 
        IOT
        &
        \textbullet~Design joint encoder for multi-modal data.\newline
        \textbullet~LLM operates as a knowledge generator.
        &
        \textbullet~One encoder and decoder can handle various types of data.\newline
        \textbullet~Achieves better coding rates and reconstruction error.
        &
        \textbullet~High computation resources due to the utilization of LLM.
        & 
        \cite{10670195}
        \\      
        \hline 
        GMAC
        & 
        IOT
        &
        \textbullet~Utilize LLM to achieve semantic alignment between observed states and natural language, and compress semantic information.
        &
        \textbullet~Improves convergence rate.\newline
        \textbullet~Enable multi-agent collaboration without communications.
        &
        \textbullet~High computation resources due to the utilization of LLM.
        & 
        \cite{10720863}
        \\      
        \hline 
        LLM-Blender 
        & 
        NLG
        &
        \textbullet~Ensemble approaches of various LLM agents for candidate ranking.
        &
        \textbullet~Ability to generate outputs better than \par the existing candidates. 
        &
        \textbullet~To achieve optimal solution, need $\mathcal{O}(n^2)$ inference times, leads to computation overheads.
        & 
        \cite{jiang-etal-2023-llm}
        \\      
        \hline 
        SOT
        & 
        NLG
        &
        \textbullet~Generate the skeleton of answer. \par
        \textbullet~Complete the contents of each skeleton in parallel.
        &
        \textbullet~Accelerate inference speed with parallel. \par
        \textbullet~Suitable for questions that require long answers (need planned structure). 
        &
        \textbullet~Answer quality evaluation is far from perfect due to limited prompt set.\par
        \textbullet~May hurt serving throughput due to parallel requests from different agents.
        & 
        \cite{ning2024skeletonofthought}
        \\
        \hline
        Meta-Prompting
        & 
        NLG
        &
        \textbullet~Construct high-level meta prompt to instruct LLMs.
        &
        \textbullet~Maintain coherent line of reasoning.\par
        \textbullet~Tapping into a varierty of expert roles.
        &
        \textbullet~Elevated cost with multiple model calls.\par
        \textbullet~Requirement for substantial scale and considerable context window.
        & 
        \cite{suzgun2024metapromptingenhancinglanguagemodels}
        \\
        
        
         
         
         
         
         
        \hline        
        MAD
        & 
        NLG
        &
        \textbullet~Two agents express their own arguments.\par
        \textbullet~A judge monitors and manages the debate.
        &
        \textbullet~Reduce bias and distorted perception.\par
        \textbullet~Encourages unlimited external feedback.
        & 
        \textbullet~Requires high computational cost due to long debate.\par
        \textbullet~LLMs struggle to maintain coherence and relevance in long scenarios.
        & 
        \cite{liang-etal-2024-encouraging} 
        \\
        \hline        
        FORD
        & 
        NLG
        &
        \textbullet~Include three-stage debate: 1) fair debate, 2) mismatched debate, 3) roundtable debate.
        &
        \textbullet~Allow LLMs to explore differences between their own understandings and the conceptualization of others via debate.
        & 
        \textbullet~Can not cover various tasks besides commonsense reasoning.\par
        \textbullet~Intensively based on the multiple choice task, which limits FORD's generalization.
        & 
        \cite{xiong-etal-2023-examining}
        \\
        
         
        
         
         
         
        \hline        
        ChatDev
        & 
        NLG
        &
        \textbullet~Employs a chat chain to break each phase into smaller subtasks, enabling multi-turn communication among agents to collaboratively develop solutions.
        &
        \textbullet~Minimizes coding halluciations, where the provided source code is missing.
        & 
        \textbullet~Without clear, detailed requirements, agents struggle to grasp task ideas.\par
        \textbullet~Automating the evaluation of general-purpose software is highly complex.\par
        \textbullet~Multiple agents require more token and times, resulting computational demands.
        & 
        \cite{qian-etal-2024-chatdev}
        \\  
         
         
         
         
        \hline        
        AgentVerse
        & 
        NLG
        &
        \textbullet~Composed of four stages: expert recruitment, collaborative decision making, action execution, evaluation.
        &
        \textbullet~Improves the generalizability of LLMs in finding uncertain.
        \textbullet~Improve the adaptability of agents.
        & 
        \textbullet~Challenges in communication among agents during the collaborative decision-making process.
        & 
        \cite{chen2024agentverse}
        \\     
        
        
        \hline        
        AgentCoord
        & 
        S\&C
        &
        \textbullet~Structured representation for coordination strategy.\par
        \textbullet~Three-stage method to transform general goal into executable strategies.
        &
        \textbullet~Streamline the representing and exploring coordination strategies.\par
        \textbullet~Minimize repetitive instances of agent engagement.
        &
        \textbullet~Only supports coordinating agents to collaborate in a plain text environment.\par
        \textbullet~Only supports static coordination strategy design.
        & 
        \cite{pan2024agentcoordvisuallyexploringcoordination}
        \\
        
        \hline
        OpenAI's Swarm & NLG & \textbullet~Routines \& Handoffs for multi-agent orchestration \par \textbullet~Lightweight framework for coordination \& execution & 
        \textbullet~Suitable for applications that require scalability \par \textbullet~Handoff mechanism allows for seamless transitions between specialized agents & \textbullet~Concern mainly with role-based protocol \& centralised/decentralized structure \par \textbullet~Not yet production-ready & See: \ref{foot:swarm}
        \\
        
        \hline 
        TE & S\&C & \textbullet~Simulate a representative sample of human participants in subject research. & \textbullet~Enables simulation of different human behaviors, and reveals consistent distortions of the simulation. & \textbullet~More human behaviors and additional LLMs needed to study to ensure the key findings. & \cite{DILLION2023597}  \\
        \hline
        AgentInstruct & S\&C & \textbullet~Generates diverse natural language data with iterative cross-agent refinement, including cultural data & \textbullet~Ables to train more capable models from generated data through tools usage, agentic capabilities, etc. & \textbullet~Requires human to hand-construct generation flows. & \cite{mitra2024agentinstruct} \\
        \hline
        SocialMind & S\&C & \textbullet~Integrates verbal, non-verbal, and social cues to generate in-situ suggestions via augmented reality glasses. & \textbullet~Designs and leverages a multi-modal, as multi-tier collaborative agent system. & \textbullet~Requires advanced edge hardware to handle complex systems. & \cite{yang2024socialmindllmbasedproactivear} \\
        \hline
        CulturePark & S\&C & \textbullet~Prompts LLM-based agents with various cultural backgrounds to simulate cross-cultural communication. & \textbullet~Generated data allow training models with less bias and democratization. & \textbullet~Still depends on LLM's knowledge of each culture, and hence limited results for low-resource cultures. & \cite{li2024cultureparkboostingcrossculturalunderstanding} \\
        \hline
        Mango & S\&C & \textbullet~Extracts high-quality knowledge from LLM-based agents through prompting on concepts and cultures. & \textbullet~Automated method allows for generating a large amount of resources. & \textbullet~Human evaluations need to be from more diverse backgrounds. & \cite{10.1145/3627673.3679768} \\
      \bottomrule
      \end{tabular}
      }
    \end{table*}    

    \subsection{5G/B6G and Industry 5.0}
    Recently, LLM has emerged to be an efficient tool to significantly improve the performance of edge networks \cite{10384606, 10638533, 10639525}. 

\begin{figure}
        \includegraphics[width=0.9\textwidth]{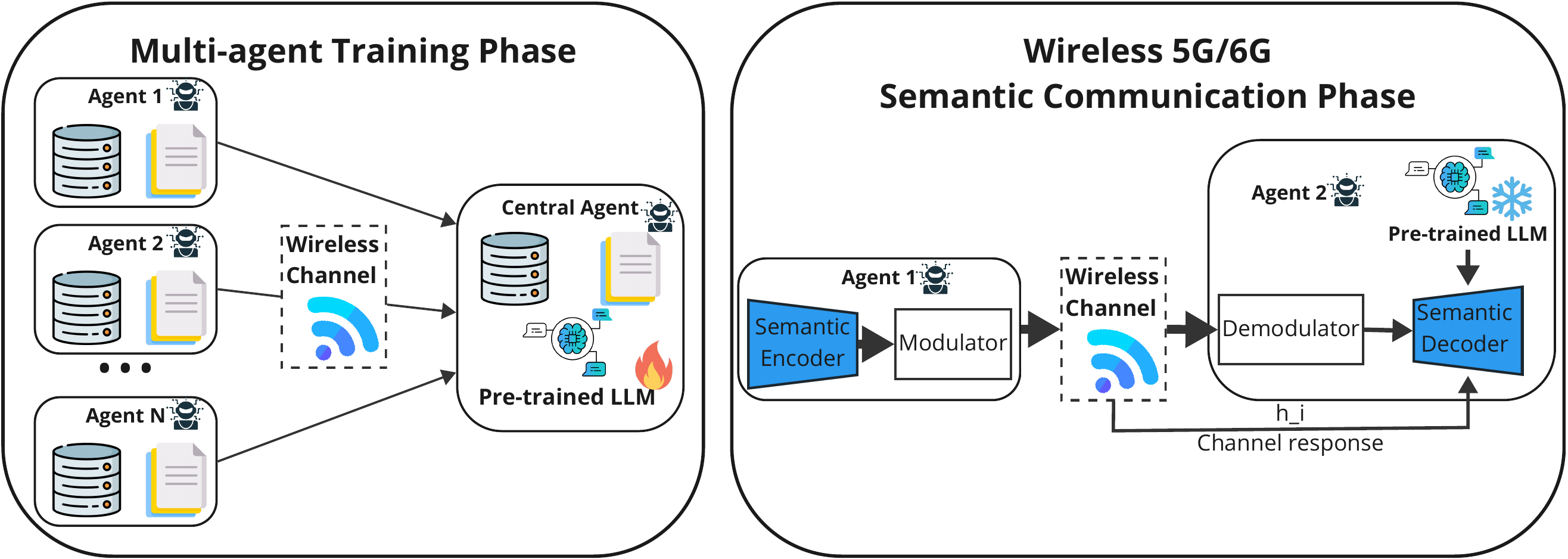}
        \caption{LLM-based MAS-enabled semantic communication system framework, leveraging LLMs directly to the physical layer coding and decoding of communication system~\cite{wang2024largelanguagemodelenabled}.}\label{fig:LLM-SC}
        \Description{}
    \end{figure}

    \textbf{5G and B6G Wireless Network.}
    LLM-SC \cite{wang2024largelanguagemodelenabled} proposed a novel framework, which utilizes LLM technology to model the semantic information of text and design a semantic communication system based on LLMs (see Fig.~\ref{fig:LLM-SC}). By using LLM to probabilistically model transmitted language sequences, LLM-SC achieves a communication paradigm that balances both semantic-level and technical-level performance. LaMoSC \cite{10531769} introduces an LLM-driven multimodal fusion semantic communication system to extend unimodal transmission and improve generalization. By leveraging the extensive external knowledge of LLMs to generate prompt text, LaMoSC overcomes the limitations of conventional semantic communication systems' knowledge bases and restricted generalization capabilities. To enhance multimodal communication, a fusion encoder is designed to integrate textual and visual features from the LLM using an attention mechanism.
    LAM-MSC \cite{10670195} presents a novel application of LLMs to enhance multimodal semantic communication frameworks. In particular, the study introduces a multimodal alignment (MMA) mechanism based on a multi-modal language model (MLM), utilizing CoDi for modality transformation. This MMA supports the synchronized generation of integrated modalities by constructing a shared multimodal space. Furthermore, to enable the comprehension of personal information, the framework incorporates a knowledge base powered by an LLM, specifically leveraging GPT-4.
    Authors in \cite{yang2024rethinkinggenerativesemanticcommunication} propose a novel framework called M2GSC. In this framework, the LLM serves as shared knowledge base, plays three critical roles, including complex task decomposition, semantic representation specification, and semantic translation and mapping. It also spawns a series of benefits such as semantic encoding standardization and semantic decoding personalization.
    GMAC \cite{10720863} introduces a data transmission strategy based on semantic information extraction to reduce the volume of data transmitted in MASs effectively. In this framework, GMAC employs LLMs to achieve semantic alignment between observed states and natural language, facilitating compressed semantic communication. This approach enhances bandwidth efficiency by extracting and compressing relevant information, thereby optimizing data transmission in multi-agent communication.
    The authors in \cite{tang2024largelanguagemodelllmassisted} propose MSADM, an end-to-end health management framework for dynamic heterogeneous networks. Using local and neighboring information, MSADM covers all stages of the health management life cycle, including anomaly detection, fault diagnosis, and mitigation. By integrating an LLM as a facilitating agent, MSADM efficiently collects and processes abnormal data, reducing diagnostic errors caused by inconsistent data representations.
    The authors in \cite{10520918} propose a novel approach that integrates LLMs with reconfigurable intelligent surfaces (RIS) to enable energy-efficient and reliable communication in the Internet of Vehicles. In this RIS system, the LLM is used to deduce optimized strategies for resource allocation and signal decoding order.
    
    \textbf{Industry 5.0.}
    The authors in \cite{Xiao_2024} propose an LLM-based IoT system using open-source LLMs deployed in a local network environment. The system includes a prompt management module, a post-processing module, and a task-specific prompts database to address concerns related to data privacy and security, system scalability, and to enhance the capabilities of the LLM through integrated prompting methods.
    The authors in \cite{10729865} propose SAGE, a smart home agent with grounded execution, which employs a scheme where a user request initiates a sequence of discrete actions controlled by an LLM. SAGE manages this process through a dynamically constructed tree of LLM prompts, which guide the agent in determining the next action, assessing the success of each action, and deciding when to terminate the process.
    The authors in \cite{10742575} present an edge-based distributed learning architecture in which a large-scale road network is divided into multiple subgraphs, with data and tasks assigned to individual RSUs. To efficiently learn from this network, they propose LSGLLM, an LLM-based method that incorporates a spatio-temporal module to capture spatio-temporal correlations. LSGLLM addresses the absence of spatio-temporal features in traditional LLMs.
    The authors in \cite{10731639} explore the integration of LLMs with the Internet of Senses technology. In this approach, an edge agent employs an LLM to generate WebXR code, enabling the visualization of corresponding 3D virtual objects on head-mounted devices and estimating multi-sensorial media data.    
    CASIT \cite{10439991} integrates LLMs into IoT systems to enhance the efficiency and intelligence of data processing and operations. By employing collective intelligence, CASIT utilizes multiple LLMs for data analysis and anomaly detection. It generates reports through a step-by-step summary and classification mechanism.
    
    \subsection{Question Answering / Natural Language Generation (QA/NLG)}

\begin{figure}[!ht]
        \centering
        \includegraphics[width=\linewidth]{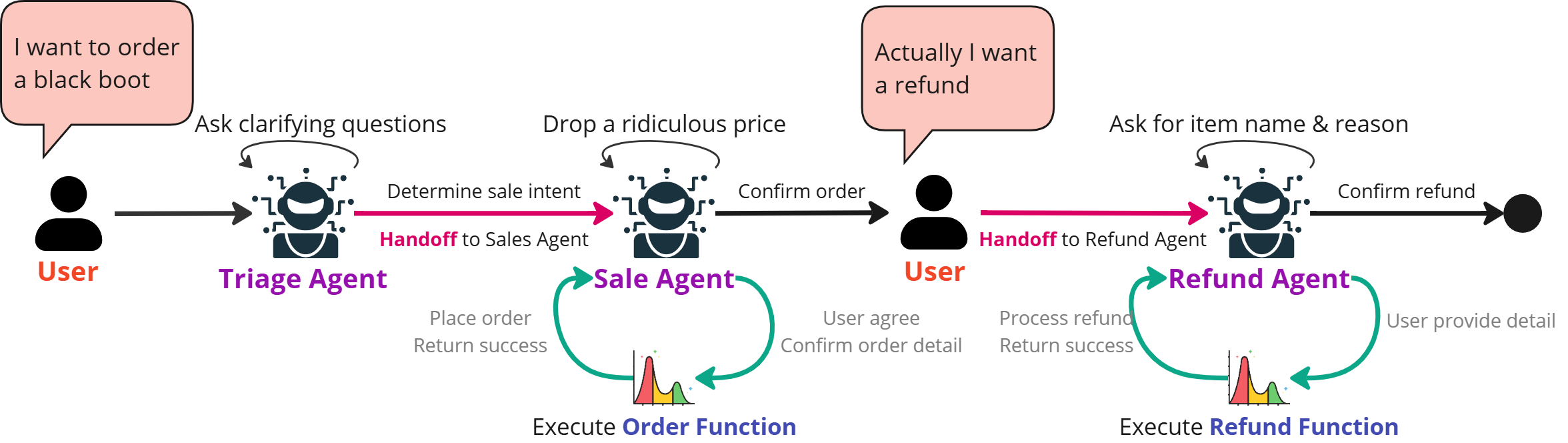}
        \caption{OpenAI's Swarm use case of customer service.}
        \label{fig:openai-swarm}
    \end{figure}

    The integration of Large Language Models (LLMs) into MASs has significantly advanced the capabilities of question answering and natural language generation. There are several prominent frameworks currently developed by leading technology companies, each employing unique mechanisms to facilitate agent collaboration in practical applications:
    
    \textbf{OpenAI's Swarm Framework}\footnote{\label{foot:swarm}Refer to OpenAI Cookbook at \url{https://cookbook.openai.com/examples/orchestrating\_agents}} : this framework introduces a novel approach to orchestrating multiple agents through the concepts of routines and handoffs. In this framework, an agent is defined as an entity that encompasses specific instructions and tools that are capable of transferring an active conversation to another agent, a process termed a "handoff." This mechanism allows for seamless transitions between agents, each specialized in particular tasks, thereby enhancing the system's overall efficiency and adaptability. Swarm's design emphasizes lightweight coordination and execution, making it suitable for scalable, real-world applications. An example with customer service focuses on sales and refunds is illustrates in the diagram \ref{fig:openai-swarm}, demonstrating the feasibility of using Swarm in pratical application.

    \textbf{Microsoft's Magentic-One System}\footnote{\url{https://www.microsoft.com/en-us/research/articles/magentic-one-a-generalist-multi-agent-system-for-solving-complex-tasks/}}: this is a generalist MAS designed to address complex tasks across diverse domains. At its core is the Orchestrator agent, which is responsible for high-level planning, progress tracking, and dynamic re-planning to recover from errors. The Orchestrator delegates specific tasks to specialized agents, such as operating a web browser, navigating local files, or writing and executing Python code. This modular architecture allows for the integration of various skills, facilitating the system's adaptability to a wide range of scenarios.
    
    \textbf{IBM's Bee Agent Framework}\footnote{\url{https://i-am-bee.github.io/bee-agent-framework/\#/}}: This open-source framework facilitates the development and deployment of scalable, multi-agent workflows. It provides a foundation for building applications where multiple AI agents, powered by LLMs such as IBM Granite and Llama 3, collaborate to achieve complex goals. The framework offers a modular design with prebuilt components for agents, tools, memory management, and instrumentation. Notably, Bee supports the serialization of agent states, enabling the pausing and resuming of complex workflows without data loss. It emphasizes modularity, extensibility, and production-level control to create sophisticated MASs for a wide range of applications, with future development aimed at enhanced multi-agent orchestration.
    
    \textbf{LangChain Agents}\footnote{\url{https://python.langchain.com/docs/tutorials/agents/}}: LangChain offers a framework for developing applications powered by language models, with a particular focus on agents. These agents are designed to interact with their environment, using tools to process information. LangChain provides a suite of tools and integrations that facilitate the creation of agents capable of complex reasoning and decision-making processes. This framework supports the development of sophisticated applications that leverage the capabilities of LLMs for advanced question answering and natural language generation tasks.
    
    These frameworks represent significant efforts in the field of multi-agent collaboration for formulating a generalized structure for building multi-agent applications, particularly in the context of question answering and natural language generation. By allowing specialized agents to work in concert, they improve the efficiency and effectiveness of AI systems, paving the way for more sophisticated and adaptable applications.

    Another trend in this area of MAS applications for QA/NLG is the introduction of novel frameworks for evaluating responses given by agents and LLMs, which reflects a reimagination of how a task should be judged - compared to the prevalently existing evaluation approach by using more capable models to give ratings, or crowd-sourced AI benchmarking from human preference \cite{chiang2024chatbot}. For example, "Agent-as-a-Judge" formulates a novel framework for evaluating agentic systems - software agents powered by LLMs - in natural language generation and question answering \cite{zhuge2024agent}. The core concept involves using agentic systems to assess other agentic systems, providing detailed feedback throughout the task-solving process, and mirroring human evaluation but at a significantly reduced cost and time. The system employs a role-based strategy where specialized agent modules (e.g., graph construction, code retrieval) operate independently in a decentralized manner with distinct functionalities, contributing to the overall evaluation. Experiments demonstrate that Agent-as-a-Judge aligns closely with human expert evaluations and surpasses the performance of traditional LLM-as-a-Judge methods, especially in complex scenarios, on the DevAI benchmark with 55 realistic AI development tasks. Another framework, "Benchmark Self-Evolving", leverages a MAS to modify existing benchmark instances by altering contexts or questions, thereby creating new, challenging instances that extend the original benchmarks \cite{wang2024benchmark}. It employs the role-based strategy, with each agent having a specific function (e.g., instance pre-filter, creator and verifier, candidate option formulator). Experiments conducted on mathematical, logical and commonsense reasoning demonstrate that the self-evolving benchmarks are more challenging than the original ones, thus offering a more accurate assessment of LLMs' true capabilities and limitations, while also addressing issues like data contamination.

    The issue of lacking data for LLM training can be alleviated by synthetic data, in which adopting MAS is considered a new approach for such task of NLG. Orca-AgentInstruct\footnote{\url{https://www.microsoft.com/en-us/research/blog/orca-agentinstruct-agentic-flows-can-be-effective-synthetic-data-generators/}} (formerly AgentInstruct), a novel agentic solution for generating high-quality synthetic data, uses a multi-agent framework to create tailored datasets from raw data sources, enabling a "generative teaching" approach for improving model capabilities in different areas \cite{mitra2024agentinstruct}. Ultilizing 3 distinct agentic flows (Content Transformation, Seed Instruction Generation, and Instruction Refinement) and decentralised structure of agents in each flow, it showed significant performance gains when used to fine-tune a Mistral 7B model, achieving improvements of up to 54\% across various benchmarks. Orca-AgentInstruct project represents a significant step towards building a synthetic data factory for model customization and continuous improvement.

    In summary, the capabilities of QA/NLG in different tasks have been improved by integrating the MAS mechanism into the process. Response evaluation in QA is now done with higher confidence, since the MAS evaluation systems resemble the process of human evaluation and now includes more dynamic evaluation with automated modified benchmarks. The NLG task of synthesizing the data is also carried out with higher-quality training data generated from the collaboration mechanism. Several notable frameworks recently introduced by big-tech companies also pave the way for the easier creation of MASs, promoting the development of such systems in practical applications. It is important to recognize that these early efforts are still in the process of being adapted and that the efficacy of applying them in practice will take time to be assessed. In addition, the integration of different types and strategies of collaboration, communication structures, and orchestration architecture also need to be considered, since most existing frameworks or systems are focusing primarily on role-based strategy and either centralized or decentralized structures.
    
        
    \subsection{Social and Cultural Domains}

\begin{figure}
            \includegraphics[width=0.72\textwidth]{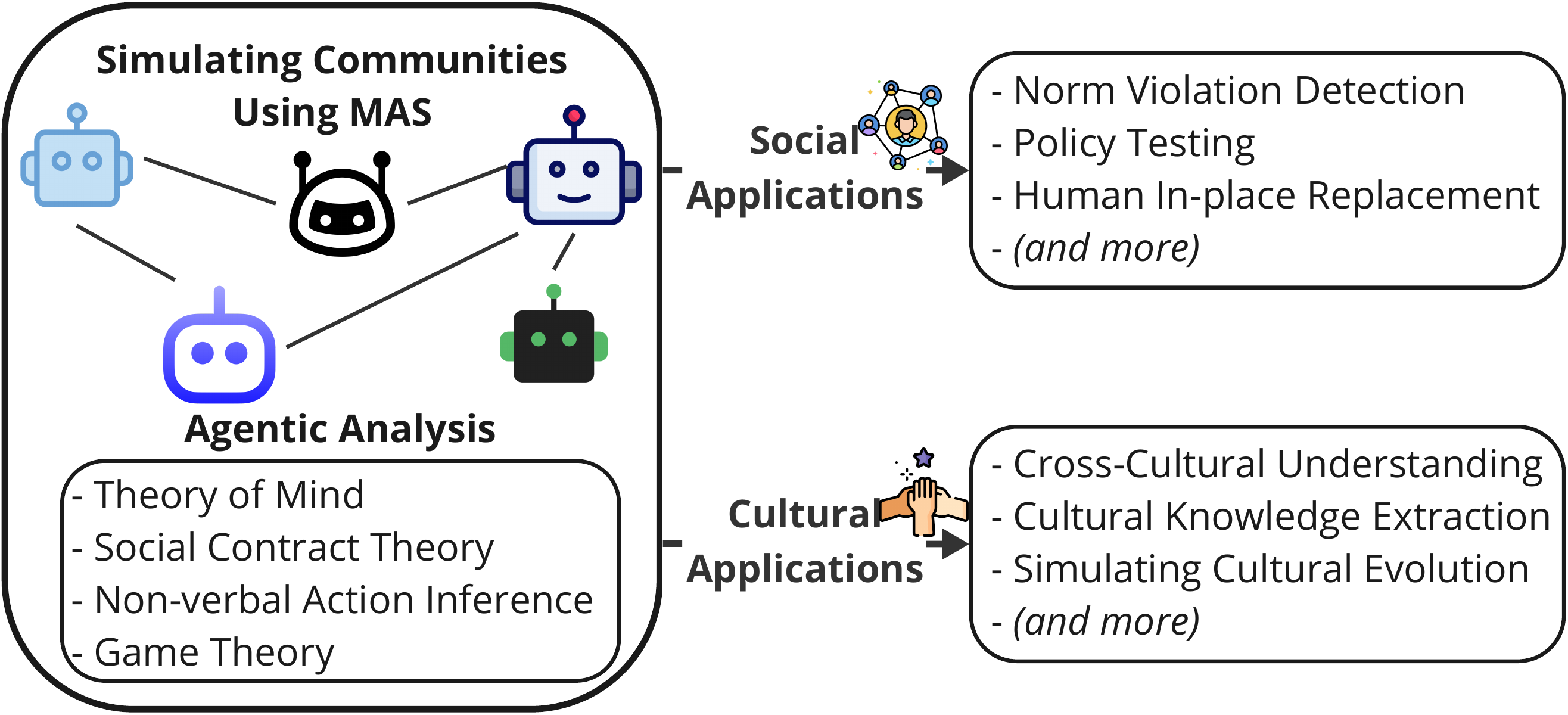}
            \caption{LLM-based multi-agent collaborative system in social \& cultural applications.}\label{fig:social_cultural_applications}
            \Description{}
        \end{figure}

        Research on LLMs and MASs has showcased the capability and applicability of these systems to simulate human behaviors, social dynamics, and cultural interactions, offering novel methodologies for understanding complex societal phenomenons, as illustrated in Fig.~\ref{fig:social_cultural_applications}. Studies such as~\cite{doi:10.1073/pnas.2314021121,10.1093/pnasnexus/pgae245} argue the potential of LLMs to enhance traditional social science methods, including survey research, online experiments, automated content analyses, and agent-based modeling. However, these studies also underscore critical limitations, such as biases in training data that lack global psychological diversity, cautioning against treating stand-alone LLMs as universal solutions. The shift from stand-alone LLMs to Multi-Agent Collaborative Systems can not only enable the analysis of LLMs in replicating individual social behavior but also provide powerful tools for exploring complex social dynamics, collaborative problem-solving, and emergent collective behaviors~\cite{Gao2024}.

        Several studies have focused on simulating social interactions through carefully designed environments, where agents are equipped with diverse prompts or LLMs tailored to specific roles. For instance,~\cite{mou2024agentsensebenchmarkingsocialintelligence} follows the definition of social interaction as a theatrical performance, with agents assuming roles (e.g., office employees or family members) driven by LLMs like GPT-4, Qwen2.5-14b, or Llama-3-8b. These roles include specific goals, such as providing and receiving feedback during office discussions. Research has shown that collaboration among LLM-based agents can elicit human-like capabilities, including conversational skills, theory of mind (reasoning about others' mental states)~\cite{li-etal-2023-theory,abdelnabi2024cooperation}, Hobbesian social contract theory (submit to authority to avoid chaos)~\cite{dai2024artificialleviathanexploringsocial}, and non-verbal action inference~\cite{Ying2024,liu2024largelanguagemodelsassume}.

        \textbf{Social Applications.} Authors in~\cite{DILLION2023597,10.5555/3618408.3618425} suggest that LLM-based agents can replace human participants in specific social science experiments, while~\cite{mitra2024agentinstruct} demonstrates their use in generating diverse natural language data with iterative cross-agent refinement. Moreover, multimodal AI systems such as those described in~\cite{yang2024socialmindllmbasedproactivear} integrate verbal, non-verbal, and social cues as input to multi-tier collaborative agents to generate in-situ suggestions via augmented reality glasses. Integrating LLM-based agents into traditional agent-based modeling~\cite{gurcan2024llmaugmentedagentbasedmodellingsocial} can enhance the realism of simulations, offering controlled environments to test social theories, including the effects of policy interventions~\cite{ansaldo2023agentspeak,zeng2024exploring} and norm violation detection~\cite{he2024normviolationdetectionmultiagent}.

        \textbf{Cultural Applications.} LLM-based MASs can represent diverse cultural perspectives, advancing cross-cultural understanding and reducing bias. For example, the CulturePark framework~\cite{li2024cultureparkboostingcrossculturalunderstanding} simulates cross-cultural interactions, with each agent embodying distinct cultural viewpoints. Similarly, Mango~\cite{10.1145/3627673.3679768} iteratively extracts high-quality cultural knowledge from LLM-based agents, providing a rich dataset for fine-tuning models to improve their ability to align with diverse cultural contexts. Another emerging area involves simulating cultural evolution within LLM populations. By modeling how cultural information is transmitted and transformed among agents, researchers gain insights into both human cultural dynamics and their influence on LLM behavior~\cite{tran2024irish,perez2024culturalevolutionpopulationslarge}. Another area of application involves simulating cultural evolution within LLM populations. By modeling how cultural information is transmitted and transformed among agents, researchers gain insights into both human cultural dynamics and their influence on LLM behavior~\cite{perez2024culturalevolutionpopulationslarge}.
        
        Despite their promise, LLMs are not perfect replicas of humans and cannot fully replicate the complexities of human social and cultural behavior. For instance,~\cite{DILLION2023597} highlights the limitations of using LLMs as human replacements in social science experiments, particularly in scenarios involving information asymmetry (unequal access to private mental states or goals)~\cite{zhou-etal-2024-real}, and in tasks requiring competition and conflict resolution~\cite{mou2024agentsensebenchmarkingsocialintelligence}. To address these challenges, consistent and standardized benchmarking approaches are necessary to evaluate the cultural and social awareness of LLM-based agents~\cite{qiu2024evaluatingculturalsocialawareness}.

\section{Open Problems \& Discussion} \label{sec:OpenProblem}
    %


    \subsection{The Road to Artificial Collective Intelligence}
        Collective intelligence is the ability of a group to perform complex tasks and solve problems collectively, often overcoming the sum of individual contributions~\cite{Leimeister2010}. With increasingly complex capabilities that mimic characteristics of living organisms, LLMs are being treated as ``digital species''. Enabling collective intelligence through collaborations among multiple LLM-based agents offers the potential for AI systems that are adaptive, efficient, and capable of addressing real-world problems. However, several open challenges must be addressed to realize this potential fully.

        \textbf{Unified Governance.} Unified governance is fundamental in enabling collective intelligence among group of LLM-based agents, including the design of coordination and planning mechanisms. Deciding which steps to take, which agents to involve, and how tasks should be distributed among them requires advanced mechanisms. Assigning specific roles or specializations to individual agents can enhance the system's overall effectiveness. Determining optimal role assignments and ensuring agents can adapt to dynamic task requirements are ongoing research areas. Moreover, governance must account for potential failures, such as miscommunication or task disruptions. Designing robust mechanisms to detect and recover from such failures is vital for ensuring the reliability and resilience of MASs. For example, introducing redundancy or fallback agents may help maintain system functionality even in adversarial scenarios.

        \textbf{Shared Decision Making.} Beyond governance, MASs must achieve coherent and accurate collective decision-making. Current LLM-based MASs commonly utilize limited decision-making methods, such as dictatorial or popular voting, which may not capture different aspects of agent preferences, or aggregating overconfidence of LLMs. Research into novel decision making approaches can enhance the diversity and fairness of collective decisions. 

        \textbf{Agent as Digital Species.} LLMs are increasingly being viewed as digital species; however, they were not originally designed for agentic applications involving collaboration and multi-participant interactions. They suffer from known limitations such as hallucinations and are susceptible to adversarial attacks. A single agent’s hallucination can be spread and reinforced by other agents, leading to minor inaccuracies into critical and cascading effects. Addressing these issues requires techniques to not only detect and correct individual errors but also to control the collaboration channels between agents. Designing LLMs specifically for collaborative environments, such as Gemini 2.0\footnote{\url{https://blog.google/products/gemini/google-gemini-ai-collection-2024/}}, represents a step toward refining these ``digital species'' for agentic systems.

        \textbf{Scalability and Resource Maintainance.} Increasing agent population poses a significant challenge in MASs. Managing resources (memory, processing time), coordination and collaboration channels among a growing number of agents introduces additional complexities, such as maintaining efficiency in agent interactions and preventing bottlenecks. Understanding the scaling laws of the behavior and performance of MASs is critical for designing architectures capable of handling large-scale collaboration.

        \textbf{Discovering/Exploring Unexpected Generalization.} Complex, emergent behaviors of collective intelligence, such as coordinated problem-solving or innovation, can arise under the right conditions, especially in generalizing to unseen domains. However, identifying and fostering these conditions is an ongoing challenge. Understanding how these generalizations emerge from the interactions of agents is key to acquiring collective intelligence.

    \subsection{Comprehensive Evaluation and Benchmarking}
        Evaluation of MASs presents challenges beyond the evaluation of individual LLMs. While there has been active research in exploring various aspects of LLMs~\cite{chang2024survey}, including their decision making capabilities and tool usage in agentic applications~\cite{patil2024gorilla,peng2024survey}, relatively few effort has been dedicated to systematically assessing the performance and behavior of LLM-based MASs~\cite{liu2024agentbench}.

        The collaborative nature of these systems introduces complexities that require a broader set of evaluation criteria. These criteria include assessing the overall system performance~\cite{zhao2024competeai}, such as reasoning capabilities, task completion rates, as well as specific system characteristics like coordination efficiency and contextual appropriateness~\cite{10.1145/3593013.3594033}. Fine-grained evaluation at the agent and collaboration levels enable root cause analysis~\cite{ji2024testingunderstandingerroneousplanning}, offering insights into individual agent behaviors, the effectiveness of collaboration channels, and the system's overall dynamics.

        Furthermore, evaluations of MASs are often conducted in narrow scenarios with different configurations, leading to inconsistent and incomparable results~\cite{davidson2024evaluating,chen-etal-2024-llmarena,chan2024chateval}. The absence of standardized evaluation protocols prevents the ability to objectively compare different systems and track progress across the field. Establishing unified, broad, and comprehensive benchmarking frameworks is vital to ensure reproducibility and consistency in evaluating MASs. Moreover, static evaluation benchmarks risk becoming lack of relevance to current real-world scenarios, leading to data leakage and overfitting~\cite{peng2024survey}. Therfore, there is a need for implementing dynamic benchmarking systems that evolve alongside technological and informational advancements.


    \subsection{Ethical Risk and Safety}
        Intrinsicly, LLMs can be harmful with hallucinated information. When deployed in MASs, these issues can propagate and amplify through agent interactions. There are two key factors behind this amplification: LLM overconfidence problem, where LLMs persistently assert the correctness of their outputs despite inaccuracies~\cite{zhang-etal-2024-exploring}, and misunderstandings that arise between LLM-based agents during collaboration. Additionally, LLMs are vulnerable to adversarial attacks, which make MASs particularly attractive targets for exploitation~\cite{shayegani2023survey}. Compromised agents in such systems can be manipulated to execute harmful or malicious behaviors. As the number of agents in LLM-based MAS increases, these risks scale proportionally, compromising the safety and reliability of communication and information exchange.

        Recent studies have also highlighted the potential for AI systems to deceive humans, raising significant concerns in the context of LLM-based multi-agent collaborative systems~\cite{meinke2024frontiermodelscapableincontext,deshpande2023anthropomorphizationaiopportunitiesrisks,Akbulut2024}. These systems, capable of simulating human societies and exhibiting human-like psychological traits, can blur the line between artificial and human behavior. Attributing human-like qualities to these systems risks fostering over-reliance, where users place trust in their capabilities. This perception can increase susceptibility to manipulation and obscure the inherent limitations of LLM-based agents~\cite{zhou-etal-2024-real}. Overlooking these limitations not only undermines informed decision-making but also introduces broader ethical concerns\footnote{\url{https://artificialintelligenceact.eu/}} regarding the responsible deployment and use of LLM-based MASs.
    
\section{Conclusion} \label{sec:Conclusion}
    Through our extensive review of the collaborative aspect of LLM-based MASs, we introduce a structured and extensible framework as an important lens to guide future research. Our framework characterizes collaboration along five key dimensions: actors, types, structures, strategies, and coordination mechanisms, providing a systematic approach to analyze and design collaborative interactions within MASs empowered by LLMs. We believe this work will inspire future research and serve as a foundational step in advancing MASs toward more intelligent and collaborative solutions.

\begin{acks}
This research work has emanated from research conducted with financial support from Science Foundation Ireland under Grant 12/RC/2289-P2 and 18/CRT/6223.
\end{acks}
\bibliographystyle{ACM-Reference-Format}
\bibliography{Reference}










\end{document}